\makeatletter

\PassOptionsToPackage{svgnames}{xcolor}

\PassOptionsToPackage{noend}{algorithmic}
\PassOptionsToPackage{noend}{algpseudocode}

\AddToHook{package/lineno/before}{\RequirePackage{amsmath}}

\AtBeginDocument{
  \@ifpackageloaded{theapa}{
    \NewDocumentCommand{\citet}{o m}{
      \IfNoValueTF{#1}
        {\citeauthor{#2} (\citeyear{#2})}
        {\citeauthor{#2} (\citeyear[#1]{#2})}
    }
    \NewDocumentCommand{\citep}{o m}{
      \IfNoValueTF{#1}
        {\cite{#2}}
        {\cite[#1]{#2}}
    }
  }{
  }
}

\AddToHook{package/algorithmic/after}{
  \renewcommand{\algorithmiccomment}[1]{\hfill \# #1}
  \def\State\STATE
  \def\If\IF
  \def\Then\THEN
  \def\Elsif\ELSIF
  \def\Else\ELSE
  \def\Endif\ENDIF
  \def\For\FOR
  \def\Forall\FORALL
  \def\Do\DO
  \def\Endfor\ENDFOR
  \def\While\WHILE
  \def\Endwhile\ENDWHILE
  \def\Repeat\REPEAT
  \def\Until\UNTIL
  \def\Return\RETURN
  \def\Require\REQUIRE
  \def\Ensure\ENSURE
  \def\Comment\COMMENT
}

\AddToHook{package/algpseudocode/after}{
  \algrenewcommand\algorithmicindent{0.7em}
  \algrenewcommand{\algorithmiccomment}[1]{\hfill \# #1}
  \algblockdefx[method]{Method}{EndMethod}[2]{\textbf{method} \textsc{#1} (#2)}{\algorithmicend}
  \ifthenelse{\equal{\ALG@noend}{t}}
   {
   \algtext*{EndMethod}
   }{}
  \algblockdefx[flet]{Flet}{EndFlet}[2]{\textbf{function} \textsc{#1} (#2)}{\algorithmicend}
  \ifthenelse{\equal{\ALG@noend}{t}}
   {
   \algtext*{EndFlet}
   }{}
  \algblockdefx[try]{Try}{EndTry}[0]{\textbf{try}}{\algorithmicend}
  \ifthenelse{\equal{\ALG@noend}{t}}
   {
   \algtext*{EndTry}
   }{}
  \algblockdefx[except]{Except}{EndExcept}[2]{\textbf{except} \textsc{#1} \textbf{as} #2}{\algorithmicend}
  \ifthenelse{\equal{\ALG@noend}{t}}
   {
   \algtext*{EndExcept}
   }{}
  \algblockdefx[finally]{Finally}{EndFinally}[0]{\textbf{finally}}{\algorithmicend}
  \ifthenelse{\equal{\ALG@noend}{t}}
   {
   \algtext*{EndFinally}
   }{}
  \def\STATE\State
  \def\IF\If
  \def\THEN\Then
  \def\ELSIF\ElsIf
  \def\ELSE\Else
  \def\ENDIF\EndIf
  \def\FOR\For
  \def\FORALL\ForAll
  \def\DO\Do
  \def\ENDFOR\EndFor
  \def\WHILE\While
  \def\ENDWHILE\EndWhile
  \def\REPEAT\Repeat
  \def\UNTIL\Until
  \def\RETURN\Return
  \def\REQUIRE\Require
  \def\ENSURE\Ensure
  \def\COMMENT\Comment
}

\makeatother

\documentclass{article}
\usepackage[position,nonatbib,preprint]{neurips_2026}
\usepackage{hyperref}

\makeatletter
\@ifpackageloaded{theapa}{
}{
 \usepackage{natbib}
}
\makeatother

\usepackage{xparse}

\usepackage{amsmath}
\usepackage{amsthm}
\usepackage{amssymb}
\usepackage{amsfonts}

\usepackage{tabularx}
\usepackage{booktabs}   
\usepackage{multirow}
\usepackage{subfigure}
\usepackage{wrapfig}

\usepackage{manfnt}             

\usepackage{xspace}             
\usepackage{relsize}            
\usepackage{adjustbox}          
\usepackage{diagbox}            
\usepackage{pdflscape}          

\usepackage{microtype}

\usepackage{graphicx}
\usepackage{url}
\usepackage{comment}            
\usepackage{xcolor}             
\usepackage{colortbl}
\usepackage{tcolorbox}

\usepackage{algorithm}
\usepackage{algpseudocode} 

\usepackage{bm}
\usepackage{alphabeta}          
\usepackage{stmaryrd}

\def\eqref#1{equation~\ref{#1}}

\def\1{\bm{1}}
\def\0{\bm{0}}

\def\rh{{\textnormal{h}}}

\def\vx{{\bm{x}}}
\def\vy{{\bm{y}}}

\DeclareMathAlphabet{\mathsfit}{\encodingdefault}{\sfdefault}{m}{sl}
\SetMathAlphabet{\mathsfit}{bold}{\encodingdefault}{\sfdefault}{bx}{n}

\def\R{{\mathbb{R}}}

\def\Z{{\mathbb{Z}}}

\DeclareMathOperator*{\argmin}{arg\,min}

\newcommand{\brackets}[1]{{\left<#1\right>}}
\newcommand{\braces}[1]{{\left\{#1\right\}}}

\newcommand{\dbrackets}[1]{{\left\llbracket#1\right\rrbracket}}

\newcommand{\then}{\therefore \qquad}

\NewDocumentCommand{\diffby}{s m O{}}{
 \IfBooleanTF{#1}
  {\frac{\partial#3}{\partial#2}}
  {\frac{d#3}{d#2}}
}

\newcommand{\satisfies}{\vDash}

\RenewDocumentCommand{\to}{o o}{
 \IfNoValueTF{#1}
  {\rightarrow}
  {\IfNoValueTF{#2}
   {\xrightarrow{#1}}
   {\xrightarrow[#2]{#1}}}
}

\NewDocumentCommand{\affect}{o o}{
 \IfNoValueTF{#1}
  {\rightsquigarrow}
  {
   \IfNoValueTF{#2}
   {\rightsquigarrow^{#1}}
   {\rightsquigarrow^{#1}_{#2}}
  }
}

\renewcommand{\then}{\Rightarrow}

\renewcommand{\iff}{\Leftrightarrow}

\newcommand{\iid}{i.i.d.\xspace}

\makeatletter

\usepackage{pdftexcmds}
\usepackage{etoolbox}
\newtoggle{dev}
\togglefalse{dev}

\ifcase\pdf@shellescape
  \or
  \toggletrue{dev}
  \or
\fi

\newcommand{\todo}[1]{\iftoggle{dev}{\red{\textbf{#1}}}{}}

\usepackage{stackengine}
\stackMath
\newcommand\tsup[2][2]{
 \def\useanchorwidth{T}
  \ifnum#1>1
    \stackon[-.5pt]{\tsup[\numexpr#1-1\relax]{#2}}{\scriptscriptstyle\sim}
  \else
    \stackon[.5pt]{#2}{\scriptscriptstyle\sim}
  \fi
}

\newcommand{\function}[1]{\textsc{#1}}

\newcommand{\mycolor}[2]{\textcolor{#1}{#2}}

\newcommand{\red}[1]{\mycolor{red}{#1}}

\def\_{\\[-0.3em]}

\newlength{\maxwidth}

\newtheorem{defi}{Definition}

\let\@myref\ref

\renewcommand{\ref}[1]{\@myref{#1}\iftoggle{dev}{\todo{(Do not use ``ref'' directly!)}}{}}

\newcommand{\refsec}[1]{Sec.\,\@myref{#1}}
\newcommand{\refseq}[1]{Sec.\,\@myref{#1}}
\newcommand{\refig}[1]{Fig.\,\@myref{#1}}
\newcommand{\reftbl}[1]{Table \@myref{#1}}
\newcommand{\refstep}[1]{Step \@myref{#1}}
\newcommand{\refalgo}[1]{Alg.\,\@myref{#1}}
\newcommand{\refchap}[1]{Chap.\,\@myref{#1}}
\newcommand{\reflst}[1]{List \@myref{#1}}
\newcommand{\refeq}[1]{Eq.\,\@myref{#1}} 

\newcommand{\reftheo}[1]{Thm.\,\@myref{#1}}
\newcommand{\refline}[1]{line\,\@myref{#1}}
\newcommand{\refdef}[1]{Def.\, \@myref{#1}}
\newcommand{\refex}[1]{Example\,\@myref{#1}}
\newcommand{\refconv}[1]{Conv.\,\@myref{#1}}
\newcommand{\reffact}[1]{Fact.\,\@myref{#1}}
\newcommand{\reflemma}[1]{Lemma.\,\@myref{#1}}
\newcommand{\refcorol}[1]{Col.\,\@myref{#1}}

\newcommand{\refsecs}[2]{Sec.\,\@myref{#1}-\@myref{#2}}
\newcommand{\refseqs}[2]{Sec.\,\@myref{#1}-\@myref{#2}}
\newcommand{\refigs}[2]{Fig.\,\@myref{#1}-\@myref{#2}}
\newcommand{\reftbls}[2]{Tables \@myref{#1}-\@myref{#2}}
\newcommand{\refsteps}[2]{Steps \@myref{#1}-\@myref{#2}}
\newcommand{\refalgos}[2]{Alg.\,\@myref{#1}-\@myref{#2}}
\newcommand{\refchaps}[2]{Chap.\,\@myref{#1}-\@myref{#2}}
\newcommand{\reflsts}[2]{Lists \@myref{#1}-\@myref{#2}}
\newcommand{\refeqs}[2]{Eq.\,\@myref{#1}-\@myref{#2}}
\newcommand{\refpages}[2]{p.\pageref{#1}-\@myref{#2}}
\newcommand{\reftheos}[2]{Thm.\,\@myref{#1}-\@myref{#2}}
\newcommand{\reflines}[2]{line\,\@myref{#1}-\@myref{#2}}
\newcommand{\refdefs}[2]{Def.\,\@myref{#1}-\@myref{#2}}
\newcommand{\refexs}[2]{Example\,\@myref{#1}-\@myref{#2}}
\newcommand{\refconvs}[2]{Conv.\,\@myref{#1}-\@myref{#2}}
\newcommand{\reffacts}[2]{Facts.\,\@myref{#1}-\@myref{#2}}
\newcommand{\reflemmas}[2]{Lemma.\,\@myref{#1}-\@myref{#2}}
\newcommand{\refcorols}[2]{Col.\,\@myref{#1}-\@myref{#2}}

\newcounter{list}[section]

\makeatother

\makeatletter

\newcommand{\pre}{\function{pre}}

\newcommand{\adde}{\function{add}}
\newcommand{\dele}{\function{del}}
\newcommand{\cost}{\function{cost}}

\def\hash{\text{\relsize{-1}\#}}
\newcommand{\ar}[1]{\hash{}#1}

\newcommand{\lsota}{state-of-the-art\xspace}  

\newcommand{\astar}{\ifmmode{A^*}\else{A$^*$}\fi\xspace}
\newcommand{\gbfs}{\ifmmode{\mathrm{GBFS}}\else{GBFS}\fi\xspace}
\NewDocumentCommand{\uct}{s}{\ifmmode{\mathrm{UCT}{\IfBooleanT{#1}{^*}}}\else{UCT{\IfBooleanT{#1}{*}}}\fi\xspace}
\NewDocumentCommand{\guct}{s}{\ifmmode{\mathrm{GUCT}{\IfBooleanT{#1}{^*}}}\else{GUCT{\IfBooleanT{#1}{*}}}\fi\xspace}

\newcommand{\newheuristic}[2]{
 \def#1{
  \relax\ifmmode
  h^\mathrm{#2}\xspace
  \else
  \text{#2}\xspace
  \fi
 }
}

\newheuristic{\lmcut}{LMcut}
\newheuristic{\mands}{M\&S}
\newheuristic{\pdb}{PDB}
\newheuristic{\ff}{FF}
\newheuristic{\ce}{CEA}
\newheuristic{\cg}{CG}
\newheuristic{\gc}{GC}
\newheuristic{\ad}{add}
\newheuristic{\hmax}{max}
\newheuristic{\lc}{LC}
\newheuristic{\blind}{blind}

\newcommand{\newlearnedheuristic}[2]{
 \def#1{
  \relax\ifmmode
  H^\mathrm{#2}\xspace
  \else
  \text{#2}\xspace
  \fi
 }
}

\newlearnedheuristic{\Hlmcut}{LMcut}
\newlearnedheuristic{\Hmands}{M\&S}
\newlearnedheuristic{\Hpdb}{PDB}
\newlearnedheuristic{\Hff}{FF}
\newlearnedheuristic{\Hce}{CEA}
\newlearnedheuristic{\Hcg}{CG}
\newlearnedheuristic{\Had}{add}
\newlearnedheuristic{\Hmax}{max}
\newlearnedheuristic{\Hlc}{LC}
\newlearnedheuristic{\Hblind}{blind}

\newcommand{\newUnitCostHeuristic}[2]{
 \def#1{
  \relax\ifmmode
  \hat{h}^\mathrm{#2}\xspace
  \else
  \text{#2}\xspace
  \fi
 }
}

\newUnitCostHeuristic{\lmcuto}{LMcut}
\newUnitCostHeuristic{\mandso}{M\&S}
\newUnitCostHeuristic{\ffo}{FF}
\newUnitCostHeuristic{\ceo}{CEA}
\newUnitCostHeuristic{\cgo}{CG}
\newUnitCostHeuristic{\ado}{add}
\newUnitCostHeuristic{\gco}{GoalCount}
\newUnitCostHeuristic{\lco}{LC}

\newcommand{\newrandomheuristic}[2]{
 \def#1{
  \ifmmode
  \rh^\mathrm{#2}\xspace
  \else
  \text{#2}\xspace
  \fi
 }
}

\newrandomheuristic{\rlmcut}{LMcut}
\newrandomheuristic{\rmands}{M\&S}
\newrandomheuristic{\rpdb}{PDB}
\newrandomheuristic{\rff}{FF}
\newrandomheuristic{\rce}{CEA}
\newrandomheuristic{\rcg}{CG}
\newrandomheuristic{\rad}{add}
\newrandomheuristic{\rhmax}{max}
\newrandomheuristic{\rlc}{LC}

\makeatother

\usepackage{etoolbox}
\usepackage{xkeyval}
\makeatletter

\def\strips@initialize{
\def\@transitiononly{0}
\def\@conditiontype{0}
\def\@usecondeffect{0}
\def\@cost{0}
\def\@useaxiom{0}
\def\@lifted{0}
\def\@track{0}
}

\define@key{strips}{transition-only}[]{\def\@transitiononly{1}}
\define@key{strips}{negative-preconditions}[]{\def\@conditiontype{1}}
\define@key{strips}{disjunctive-preconditions}[]{\def\@conditiontype{2}}
\define@key{strips}{conditional-effects}[]{\def\@usecondeffect{1}}
\define@key{strips}{action-costs}[]{\ifnumcomp{\@cost}{<}{1}{\def\@cost{1}}{}}
\define@key{strips}{unitcost}[]{\ifnumcomp{\@cost}{<}{2}{\def\@cost{2}}{}}
\define@key{strips}{derived-predicates}[]{\def\@useaxiom{1}}
\define@key{strips}{lifted}[]{\def\@lifted{1}}
\define@key{strips}{optimal}[]{\ifnumcomp{\@track}{<}{1}{\def\@track{1}}{}}
\define@key{strips}{satisficing}[]{\ifnumcomp{\@track}{<}{2}{\def\@track{2}}{}}
\define@key{strips}{agile}[]{\ifnumcomp{\@track}{<}{3}{\def\@track{3}}{}}

\def\conditionset{
\if\@useaxiom0
P
\else
P\cup P_X
\fi
}

\let\satisfies@orig\satisfies

\def\satisfies{
\if\@conditiontype0
\supseteq
\else
\satisfies@orig
\fi
}

\def\condition{
\if\@conditiontype0
\conditionset
\else
\mathcal{F}(\conditionset)
\fi
}

\def\ga{
\if\@lifted0
a
\else
a^{\dagger}
\fi
}

\def\applyformula{
\if\@usecondeffect0
(s \setminus \dele(a)) \cup \adde(a)
\else
(s
 \setminus \braces{e \mid (c \triangleright e) \in \dele(\ga), c\satisfies s})
 \cup      \braces{e \mid (c \triangleright e) \in \adde(\ga), c\satisfies s}
\fi
}

\NewDocumentCommand{\strips}{O{}}{
\strips@initialize
\setkeys{strips}{#1}
\if\@lifted1
  \strips@propositional\par
  \strips@lifted
\else
  \strips@propositional
\fi
}

\newcommand{\strips@propositional}{
\if\@conditiontype1
Given a set of propositional variables $V$,
let $\mathcal{F}(V)$ be a propositional formula consisting of $V$ and
logical operations $\braces{\land,\lnot}$.
\fi
\if\@conditiontype2
Given a set of propositional variables $V$,
let $\mathcal{F}(V)$ be a propositional formula consisting of $V$ and
logical operations $\braces{\land,\lor,\lnot}$.
\fi
\if\@useaxiom0
We define a propositional STRIPS Planning problem
as a 4-tuple $\brackets{P,A,I,G}$
where
 $P$ is a set of propositional variables,
 $A$ is a set of actions,
 $I\subseteq P$ is the initial state, and
 $G\subseteq \conditionset$ is a goal condition.
\else
We define a propositional STRIPS Planning problem
as a 6-tuple $\brackets{P,A,X,P_X,I,G}$
where
 $P$ is a set of propositions,
 $A$ is a set of actions,
 $X$ is a set of axioms,
 $P_X$ is a set of derived propositions ($P\cap P_X=\emptyset$),
 $I\subseteq P$ is the initial state, and
 $G\subseteq \conditionset$ is a goal condition.
\fi
\ifnumcomp{\@transitiononly}{>}{0}{
We omit the details of action applications as they are not important in this paper.
It suffices to say an action $a\in A$ transitions from a state $s\subseteq P$ to a successor $s'=a(s)\subseteq P$.
}{
\ifnumcomp{\@cost}{<}{1}{
Each action $a\in A$ is a 3-tuple $\brackets{\pre(a),\adde(a),\dele(a)}$ where
}{
Each action $a\in A$ is a 4-tuple $\brackets{\pre(a),\adde(a),\dele(a),\cost(a)}$ where
$\cost(a) \in \Z^{0+}$ is a cost\ifnumcomp{\@cost}{=}{2}{ (We assume unit-cost: $\forall a\in A; \cost(a)=1$)}{},
}
$\pre(a) \subseteq \condition$ is a precondition and
\if\@usecondeffect0
$\adde(a), \dele(a)\subseteq P$ are the add-effects and delete-effects.
\else
$\adde(a), \dele(a)$ are the add-effects and delete-effects.
Each effect is denoted as $c \triangleright e$ where
$c \in \condition$ is an \emph{effect condition} and
$e \in P$.
\fi
\if\@useaxiom1
The set of axioms $X$ consists of clauses $f \Rightarrow p$ where
$f \in \condition$ is a body and $p \in P_X$ is a head.
\fi
A state $s\subseteq \conditionset$ is a set of true propositions
(all of $P\setminus s$ is false),
an action $a$ is \emph{applicable} when $s \satisfies \pre(a)$ (read: $s$ \emph{satisfies} $\pre(a)$),
and applying action $a$ to $s$ yields a new successor state
\if\@useaxiom0
$a(s) = \applyformula$.
\else
$a(s)$.
To compute $a(s)$, we first obtain a non-derived state
$s' \gets \applyformula \setminus P_X $.
Then we perform a fix-point calculation
$s' \gets s' \cup \braces{p \in P_X \mid (f\Rightarrow p)\in X \land s \satisfies f}$.
\fi
\par
}                               
The task of classical planning is to find a sequence of actions called a \emph{plan} $(\ga_1,\cdots,\ga_n)$
where, for $1\leq t\leq n$,
 $s_0=I$,
 \ifnumcomp{\@transitiononly}{>}{0}{}{$s_t\satisfies \pre(a_{t+1})$,}
 $s_{t+1}=a_{t+1}(s_t)$,
 and $s_n\satisfies G$.
\ifnumcomp{\@track}{>}{0}{
 A plan is \emph{optimal} if
 \ifnumcomp{\@cost}{<}{1}{
   there is no shorter plan.
 }{
   there is no plan with a lower cost $\sum_t \cost(a_t)$.
 }
 \ifnumcomp{\@track}{>}{1}{
   A plan is otherwise called \emph{satisficing}.
   \ifnumcomp{\@track}{>}{2}{
     In particular, a problem setting that completely ignores the solution quality is called \emph{agile} setting,
     while \emph{satisficing} setting implies that the solver still attempts to find a
     \ifnumcomp{\@cost}{<}{1}{shorter}{cheaper}
     plan.
     This paper focuses on the \emph{agile} setting.
   }{
     This paper focuses on the \emph{satisficing} setting.
   }
 }{}
}{}
}

\newcommand{\strips@lifted}{
In \emph{Lifted STRIPS}, each propositional variable is an \emph{instantiation}/\emph{grounding} of
a first-order logic predicate.
Each predicate $p(x_1,\ldots,x_{\ar{p}})$ is parameterized by a list of parameters/variables/arguments $X=(x_1,\ldots,x_{\ar{p}})$,
where $\ar{p}$ is an \emph{arity} of $p$.
A proposition is obtained by substituting each $x_i$ with an \emph{object} in a set $O$.
Each $p$ therefore has $O^{\ar{p}}$ instantiations.
Similarly, each action $a\in A$ is now called a \emph{ground action},
which is an instantiation of a \emph{lifted action} $a(x_1,\ldots,x_{\ar{p}})$ parameterized by $\ar{a}$ parameters.
A ground action is obtained by substituting the arguments as well as
the parameters used in the preconditions and the effects.
}

\makeatother

\usepackage{etoolbox}
\usepackage{xkeyval}
\makeatletter

\long\def\addto#1#2#3{
  \ifinlist{#3}{#1}{
  }{
    \listadd{#1}{#3}
    \ifdefempty#2{
     \expandafter\def\expandafter#2\expandafter{#2#3}
    }{
     \expandafter\def\expandafter#2\expandafter{#2,#3}
    }
  }
}

\define@key{heuristics}{ff}[1]{
 \def\heuristics@ff{#1}
 \ifnumcomp{\heuristics@ff}{>}{0}{
  \addto{\heuristiclist}{\heuristicstr}{\ff}
  \addto{\heuristiccitelist}{\heuristiccitestr}{hoffmann01}
 }{}
}
\define@key{heuristics}{ad}[1]{
 \def\heuristics@ad{#1}
 \ifnumcomp{\heuristics@ad}{>}{0}{
  \addto{\heuristiclist}{\heuristicstr}{\ad}
  \addto{\heuristiccitelist}{\heuristiccitestr}{bonet2001planning}
 }{}
}
\define@key{heuristics}{hmax}[1]{
 \def\heuristics@hmax{#1}
 \ifnumcomp{\heuristics@hmax}{>}{0}{
  \addto{\heuristiclist}{\heuristicstr}{\hmax}
  \addto{\heuristiccitelist}{\heuristiccitestr}{bonet2001planning}
 }{}
}
\define@key{heuristics}{gc}[1]{
 \def\heuristics@gc{#1}
 \ifnumcomp{\heuristics@gc}{>}{0}{
  \addto{\heuristiclist}{\heuristicstr}{\gc}
  \addto{\heuristiccitelist}{\heuristiccitestr}{FikesHN72}
 }{}
}
\define@key{heuristics}{cea}[1]{
 \def\heuristics@cea{#1}
 \ifnumcomp{\heuristics@cea}{>}{0}{
  \addto{\heuristiclist}{\heuristicstr}{\ce}
  \addto{\heuristiccitelist}{\heuristiccitestr}{helmert2008unifying}
 }{}
}
\define@key{heuristics}{cg}[1]{
 \def\heuristics@cg{#1}
 \ifnumcomp{\heuristics@cg}{>}{0}{
  \addto{\heuristiclist}{\heuristicstr}{\cg}
  \addto{\heuristiccitelist}{\heuristiccitestr}{Helmert04}
 }{}
}

\define@key{heuristics}{simplified}[1]{\def\heuristics@simplified{#1}}

\define@key{heuristics}{relaxation}[1]{\def\heuristics@relaxation{#1}}

\define@key{heuristics}{perfect}[1]{\ifnumcomp{\heuristics@properties}{<}{#1}{\def\heuristics@properties{#1}}{}}
\define@key{heuristics}{admissible}[2]{\ifnumcomp{\heuristics@properties}{<}{#1}{\def\heuristics@properties{#1}}{}}
\define@key{heuristics}{inadmissible}[3]{\ifnumcomp{\heuristics@properties}{<}{#1}{\def\heuristics@properties{#1}}{}}
\define@key{heuristics}{perfectsat}[4]{\ifnumcomp{\heuristics@properties}{<}{#1}{\def\heuristics@properties{#1}}{}}
\define@key{heuristics}{tdom}[5]{\ifnumcomp{\heuristics@properties}{<}{#1}{\def\heuristics@properties{#1}}{}}

\define@key{heuristics}{expansion}[1]{\def\heuristics@expansion{#1}}

\NewDocumentCommand{\heuristics}{O{}}{
\def\heuristics@ff{0}
\def\heuristics@ad{0}
\def\heuristics@hmax{0}
\def\heuristics@gc{0}
\def\heuristics@cea{0}
\def\heuristics@cg{0}
\def\heuristics@relaxation{0}
\def\heuristics@simplified{0}
\def\heuristics@properties{0}
\def\heuristics@expansion{0}
\def\heuristiclist{}
\def\heuristiccitelist{}
\def\heuristicstr{}
\def\heuristiccitestr{}
\setkeys{heuristics}{#1}
\ifnumcomp{\heuristics@simplified}{>}{0}{
A domain-independent heuristic function $h(s)$
returns an estimate of the cumulative cost from a state $s$ to one of the goal states (states that satisfy $G$).
}{
Given a problem $\brackets{P,A,I,G}$ and a state $s$,
a domain-independent heuristic function $h(s, \brackets{P,A,I,G})$
returns an estimate of the cumulative cost from $s$ to one of the goal states (states that satisfy $G$),
typically through a symbolic, non-statistical means including problem relaxation and abstraction.
It is often abbreviated as $h(s)$ or $h(s,G)$.
}
\ifdefempty\heuristiclist{}{
Notable \lsota functions that appear in this paper includes
$\heuristicstr$ \citep{\heuristiccitestr}.
}
\ifnumcomp{\heuristics@properties}{<}{1}{}{
  Often, the true optimal cost from a state $s$ is called
  the \emph{perfect heuristics} $h^*(s)$ \citep{helmert2008good}.
  \ifnumcomp{\heuristics@properties}{<}{2}{}{
    \emph{Admissible} heuristics are those which never overestimate $h^*$,
    i.e., $\forall s; h(s)\leq h^*(s)$.
    Optimizing algorithms like \astar \citep{hart1968formal} are
    guaranteed to find the optimal solutions with such heuristics.
    \ifnumcomp{\heuristics@expansion}{<}{1}{}{
      Moreover, \astar is the optimal expansion algorithm, i,e.,
      expands the fewest nodes among all algorithms under the same admissible $h$.
    }
    \ifnumcomp{\heuristics@properties}{<}{3}{}{
      Otherwise they are called \emph{inadmissible} heuristics,
      and are typically combined with satisficing/agile algorithms like GBFS \citep{doran1966experiments,bonet2001planning}.
      \ifnumcomp{\heuristics@properties}{<}{4}{}{
        Furthermore, heuristics that preserve the same ordering as $h^*$ are called
        \emph{perfect satisficing heuristics} $h^\leq$ \citep{kuroiwa2022biased},
        i.e., $\forall s,t; h^\leq(s)\leq h^\leq(t)\then h^*(s)\leq h^*(t)$.
        \ifnumcomp{\heuristics@expansion}{<}{1}{}{
          GBFS is the optimal expansion algorithm under $h^\leq$.
        }
        \ifnumcomp{\heuristics@properties}{<}{5}{}{
          Given a monotonic \emph{inflation} function $t:\R^{0+}\to\R^{0+}$
          s.t. $\forall x; t(x)\geq x$ and $\forall x,y; x\geq y \then t(x)\geq t(y)$,
          heuristics that preserve the same ordering as $h^*$ when inflated are called
          \emph{$t$-dominating heuristics},
          i.e., $\forall s,t; t(h(s))\leq h(t)\then h^*(s)\leq h^*(t)$.
        }
      }
    }
  }
}

\if\heuristics@relaxation1
A significant class of heuristics is called delete relaxation heuristics,
which solve a relaxed problem which does not contain delete effects,
and then returns the cost of the solution of the relaxed problem as an output.
The cost of the optimal solution of a delete relaxed planning problem from a state $s$ is
denoted by $h^+(s)$, but this is too expensive to compute in practice (NP-complete) \citep{bylander1996}.
Therefore, practical heuristics typically try to obtain its further relaxations
that can be computed in polynomial time.
\fi

\ifnumcomp{\heuristics@hmax}{>}{1}{
Max heuristics $\hmax$ \citep{bonet2001planning} is recursively defined as follows:
\begin{align}
 \hmax(s,G) = \max_{p\in G}
 \left\{
  \begin{array}{l}
   0\ \text{if}\ p\in s.\ \text{Otherwise,}\\
   \min_{\braces{a\in A\mid p\in\adde(a)}} \\
    \quad \left[\cost(a)+\ad(s, \pre(a))\right].
  \end{array}
 \right.
\end{align}
}{}

\ifnumcomp{\heuristics@ad}{>}{1}{
Additive heuristics $\ad$ \citep{bonet2001planning} is recursively defined as follows:
\begin{align}
 \ad(s,G) = \sum_{p\in G}
 \left\{
  \begin{array}{l}
   0\ \text{if}\ p\in s.\ \text{Otherwise,}\\
   \min_{\braces{a\in A\mid p\in\adde(a)}} \\
    \quad \left[\cost(a)+\ad(s, \pre(a))\right].
  \end{array}
 \right.
\end{align}
}{}

\ifnumcomp{\heuristics@ff}{>}{1}{
FF heuristics $\ff$ \citep{hoffmann01} is defined based on another heuristics $h$, such as $h=\ad$, as a subprocedure.
For each proposition $p$,
the action $a$ that adds $p$ with the minimal $\cost(a)+h(s, \pre(a))$
is conceptually ``the cheapest action that achieves a subgoal $p$'',
called the \emph{cheapest achiever} / \emph{best supporter} $\text{bs}(p,s,h)$ of $p$.
Using this, $\ff$ is defined as the sum of actions in a relaxed plan $\Pi^+$ constructed as follows:
\begin{align}
 \ff(s,G,h) &= \sum_{a\in \Pi^+(s,G,h)} \cost(a)\\
 \Pi^+(s,G,h) &= \bigcup_{p\in G}
 \left\{
  \begin{array}{l}
   \emptyset\ \text{if}\ p\in s.\ \text{Otherwise,}\\
   \braces{a} \cup \Pi^+(s,\pre(a)) \\
   \qquad \text{where}\ a=\text{bs}(p,s,h).
  \end{array}
 \right.\\
 \text{bs}(p,s,h)&=\argmin_{\braces{a\in A\mid p\in \adde(a)}} \left[\cost(a)+h(s, \pre(a))\right].
\end{align}
\ifnumcomp{\heuristics@ff}{>}{2}{
  In practice, $\ff$ can be implemented in several ways, each producing different values
  due to the tie-breaking difference in the $\argmin$ in $\text{bs}(p,s,h)$.
  \citet{kuroiwa2019case} showed that Graphplan-based implementation yields the best planner performance
  due to the combination of low-level efficiency and heuristic accuracy.
}{}
}{}

\ifnumcomp{\heuristics@gc}{>}{1}{
Goal Count heuristics $\gc$ is a simple heuristic proposed in \citep{FikesHN72}
that counts the number of propositions that are not satisfied yet.
$\brackets{\text{condition}}$ is a cronecker's delta / indicator function that returns 1 when the condition is satisfied.
\begin{align}
 \gc(s,G) &= \sum_{p\in G} \dbrackets{p\not \in s}.
\end{align}
}{}
}

\makeatother

\iftoggle{dev}{
  \IfFileExists{\jobname.needpyg}{
    \usepackage[finalizecache,cachedir=.]{minted}
  }{
    \usepackage[cachedir=.]{minted}
  }
}{
  \usepackage[frozencache,cachedir=.]{minted}
}

\allowdisplaybreaks

\title{Verbalized Algorithms: \\Classical Algorithms are All You Need (Mostly)}
\author{
  Supriya Lall \\
  MIT CSAIL \\
  MIT-IBM Watson AI Lab \\
  \texttt{supriya@mit.edu} \\
  \And
  Christian Farrell \\
  Marist University \\
  IBM Infrastructure \\
  \texttt{christian.farrell@ibm.com} \\
  \And
  Hari Pathanjaly \\
  UC Irvine \\
  IBM Infrastructure \\
  \texttt{hari.pathanjaly@ibm.com} \\
  \And
  Marko Pavic \\
  Marist University \\
  IBM Infrastructure \\
  \texttt{marko.pavic@ibm.com} \\
  \And
  Sarvesh Chezhian \\
  IBM Infrastructure \\
  \texttt{schez@ibm.com} \\
  \And
  Masataro Asai \\
  MIT-IBM Watson AI Lab \\
  IBM Research Cambridge, USA \\
  \texttt{masataro.asai@ibm.com} \\
}

\begin{document}

\maketitle

\begin{abstract}
Reasoning is a fundamentally algorithmic task.
Yet current work on LLM-based reasoning relies on free-form generation
whose theoretical guarantees (soundness, completeness, complexity, optimality)
remain poorly understood.
We argue that we should not treat them as general-purpose reasoners, and
as an alternative,
we propose a paradigm we call \emph{verbalized algorithms} (VAs),
which combines LLMs and various algorithms 
with established guarantees.
Instead of betting on LLM's ability to solve a reasoning task,
VAs limit their scope by
decomposing the task down to simple elementary operations on strings
that they can answer reliably.
For example, sorting a list of natural language strings
could be done by using an LLM as a binary comparison oracle
in a parallel or approximate sorting algorithm.
We push the accuracy-runtime Pareto front with
\emph{verbalized maximum}, \emph{sorting}, \emph{clustering}, and \emph{submodular maximization},
for numerical reasoning, topic clustering, Wi-Fi access point optimization, and multi-hop Q\&A RAG task.
These results suggest
improving LLM-based reasoning through standard algorithmic analysis
is a feasible and better grounded research direction.

\end{abstract}

\section{Introduction}

\begin{figure*}[tb]
 \includegraphics[width=\linewidth]{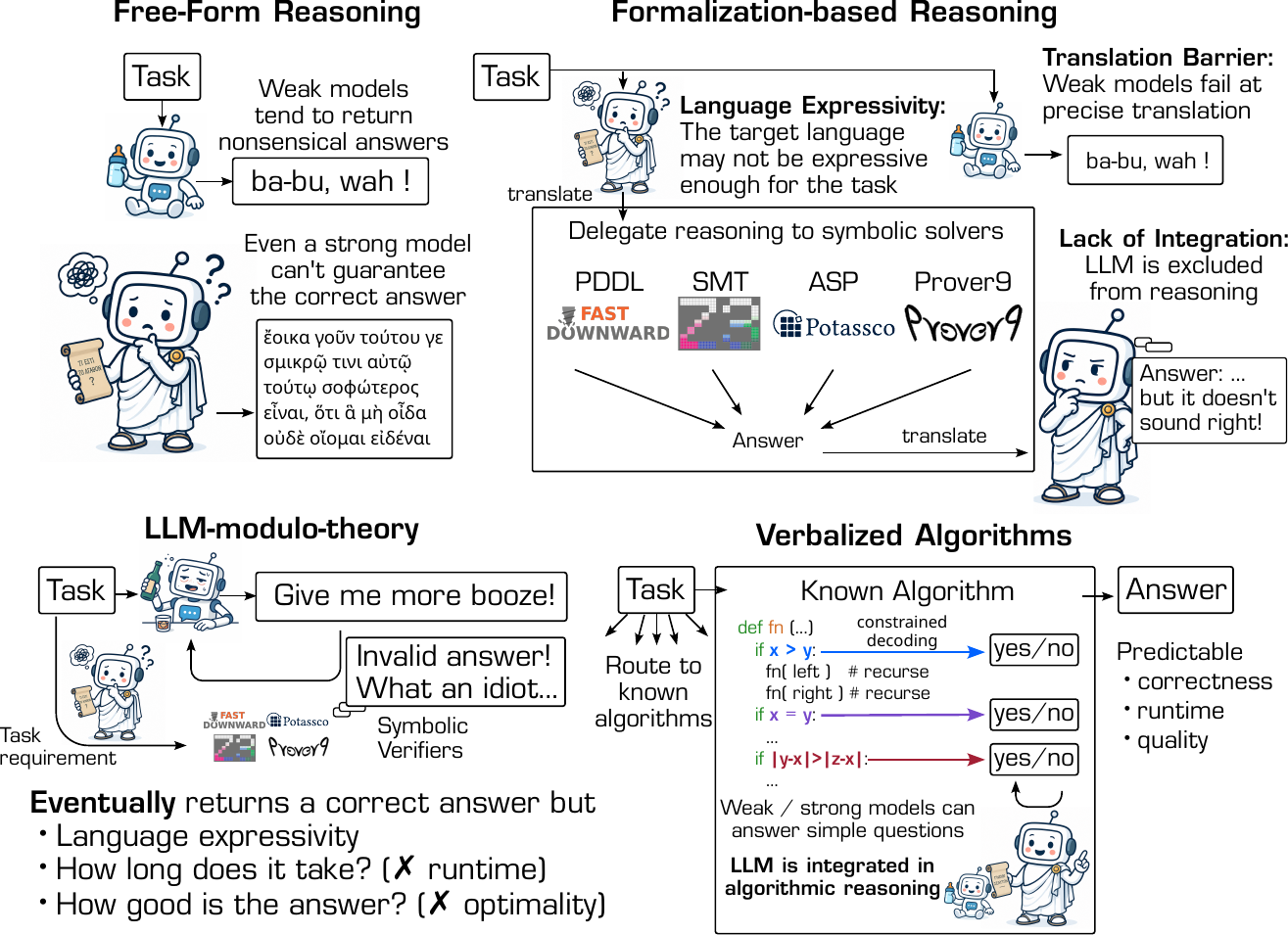}
 \caption{Conceptual comparison of various reasoning approaches}
 \label{fig:concept}
\end{figure*}

Imagine receiving thousands of customer complaints and having to
quickly sort them by how urgently they need support.
Feeding such a large set of input strings blindly into a transformer-based large language model (LLM)
is ineffective and impractical due to various limitations. Examples of such limitations include
limited context lengths from hardware limitation (GPU memory limit from $O(n^2)$ attentions),
sequential runtime for processing a long input,
deteriorating accuracy from the longer context,
and its limited ability to perform the sorting task itself.
LLMs provide no guarantees on the correctness of their outputs for computational reasoning tasks,
including but not limited to sorting,
beyond anecdotal, empirical observations that they \emph{sometimes} do.
Despite recent improvements in Chain-of-Thoughts \citep{wei2022chain} and think tags \citep{guo2025deepseek},
their behavior is fundamentally an approximation of formal reasoning, and thus contains hallucinations.

We argue that
most reasoning tasks are informal representations of formal tasks
for which \textbf{an efficient and well-analyzed classical algorithm already exists in the literature},
thus training a model hoping that it may learn to solve it by itself
is often a reinvention of the wheel.
For example, for sequential comparison sort, $O(n \log n)$ algorithms are asymptotically optimal, thus
no algorithm can improve over it anyways --- \textbf{then why don't we just use them?}

In addition to hardware limitations and long context issues,
LLM reasoning usually does not follow a formally specified algorithm.
Even if it does, the algorithm may be unsound, incomplete,
and lack formal guarantees on time/space/optimality.
Even if we discount these issues, 
autoregressive LLMs fundamentally operate sequentially.
One could parallelize token generations (e.g., beam search or MCTS),
but such task-agnostic parallelization lacks sophisticated coordinations
in dedicated parallel algorithms that exploit task characteristics.

Traditional disciplines of computer science instead approach computational tasks with \emph{task decomposition}.
For example, a merge sort decomposes a sorting task into smaller sorting tasks
and combines the results, where, at the limit, there is an elementary binary comparison over
individual pairs.
These traditional algorithms, given an accurate oracle, guarantee the correctness of their outputs
and the average/worst-case runtime/memory.
The opaqueness of LLM's reasoning prevents such an analysis.

In this work, we propose an abstract framework called ``verbalized algorithms''.
Given an informal representation of a formal task,
a verbalized algorithm replaces some atomic operations in a classical algorithm
with LLM-driven text-based oracles while \emph{retaining the high-level control flow},
thereby largely retaining the theoretical guarantees.
Our work is heavily influenced by
\emph{Algorithm with Predictions} \citep[ALPS]{mitzenmacher2022algorithms} literature,
which achieved significant progress in
augmenting algorithms with data-driven oracles.
There are certainly other work that replaces part of algorithms with LLM oracles,
including successor function and goal condition in planning \citep{michael2024tos} and
collision checking in STPR \citep{djuhera2025don},
yet we believe ours is the first general proposal for the reductionist approach
toward LLM reasoning.

We perform four case studies with increasing difficulty and non-triviality:
Maximum, sorting, clustering, and submodular maximization.
Our implementations are built on top of the offline asynchronous engine of SGLang \citep{sglang},
and all experiments are repeated 5 times with different seeds.

\textbf{Contributions.}
VAs have several key advantages:
(1)
It is more accurate than classical non-LLM methods
or more recent embedding-based or other LLM-based methods,
and is also scalable with data parallelism.
(2)
The reasoning trace of VAs is interpretable, as we have a full understanding of
its high-level algorithm and can focus on individual decision points.
(3)
VAs provide some degrees of theoretical guarantees.
The key point is not these guarantees themselves
but the fact that they \emph{transfer directly from the existing literature} even though we are using LLMs,
which massively helps reducing the chance of reinventing the wheels.
(4)
VAs work well with smaller, efficient reasoning models.
(5)
VAs are more accessible than some non-VA methods which require model training, and helps AI democratization.
Such training is typically beyond the reach of casual LLMs users,
while VAs do not require additional training.
Fine-tuning a model for VA is possible, and is a promising future work.

\textbf{Limitations.}
The main limitation of VAs is the resource intensity from the large constant factor involved in each query.
This, however,
is not a real limitation in the long term, as the progress in silicon manufacturing makes more powerful GPUs cheaper.
Moreover, VAs are designed to tackle more difficult tasks, potentially combinatorial ones,
that naturally require more precision and a more careful algorithmic procedure to address compounding errors.
Lastly,
VAs' accuracy-runtime Pareto front is on par or better than with existing approaches by using a smaller model, as seen later.

Another limitation is the cost of writing each algorithm by hand.
However, coding agents significantly lowered the bar recently,
as they often know even the most sophisticated algorithms in the literature.
Agentic workflows involving subagents writing their own VAs would accelerate this direction.

\textbf{Alternative Views.}
Another set of approach to similar tasks is what we call ``formalization-based approaches''.
Leveraging the translation capability of LLMs,
it converts an informal task representation into a formal, symbolic representation,
solves it with a specialized combinatorial solver,
and then converts the solution back to the informal, natural language representation.
Existing work on this approach includes LLM-based generation of
PDDL \citep{McDermott00} for classical planning \citep{xie2020translating,guan2023leveraging,li2024naruto,fine2024leveraging,michael2024llmpddl},
Prover9 theorem proving input \citep{olausson2023linc}, or
SMT input \citep{hao2024planning}.
The issue with this approach is twofold:
First, it requires grammar-following capability for a complex target formal language.
This could be error-prone and time-consuming, and/or
the grammar class (context free/sensitive) of the target language could exceed
the scope of constraint-decoding libraries \citep{dong2025xgrammar}.
Second,
the target language limits the scope of the tasks that can be solved.
For example, an agent that converts the task into PDDL cannot handle
a probabilistic, temporal, or partially-observable task.
VAs, on the other hand, could handle them robustly
by applying informal common sense reasoning to atomic operations.
We see the evidence of this claim in the sorting section.

LLM-Modulo framework \citep{kambhampati2024position}
views an LLM as an ``idea generator'' and combines it with
rejection sampling / feedback loop with symbolic verifiers
to improve the accuracy of LLM-based reasoning, particularly in planning tasks.
While it shares the concerns about the soundness of LLM-based reasoning with VAs,
it has three key issues.
First, in the general LLM-based natural language reasoning,
there is no guarantee that such a verifier would exist because
the target application could be highly informal (e.g., related to human emotion and ethics),
which makes it challenging to formally / objectively / symbolically define any correctness.
VAs do not have this issue --
when the algorithm contains halting/backtracking with a verifier,
the verifier could be replaced by LLM-based oracles to accommodate informal tasks.
Second, it focuses on decision problems
and there is little considerations for optimization problems.
Algorithms for optimization generally rely on ranking/comparison (discussed extensively in this paper)
and datastructures around it (e.g., priority queues).
Third,
due its reliance on symbolic verifiers,
it is also limited by the expressivity of the target language,
even if the verifier exists and the translation to the language is possible.

Another view may criticize that
each VA is specialized to some task and does not offer ``general intelligence''.
However, it can be largely resolved by tool-use agents \citep{schick2023toolformer}
which are already widely adopted by the industry.
The agent can act as a router to VA implementations.

\section{Verbalized Algorithms}

While whether LLMs can reason or not is up for both philosophical and technical debates \citep{shojaeeillusion,hazra2024can},
it is more widely accepted that they can solve some much simpler tasks
such as multi-choice question answering with high fidelity.
From this observation,
we propose \emph{verbalized algorithms} (VAs),
a reasoning strategy that limits the scope of LLMs to simple tasks.

To exhibit a trait typically useful for user interaction (e.g. instruction following),
LLMs go through a series of parameter-efficient fine-tuning (PEFT) steps after pretraining on a large corpus.
For example, in Direct Preference Optimization \citep[DPO]{rafailov2023direct},
the input is a pair of model outputs in which human users have indicated their preference over another
and the model is fine-tuned to follow that preference.
For a prompt $x$ and an answer pair $y_1$ and $y_2$, denote by $y_1 \succ y_2 | x$ that $y_1$ is preferred over $y_2$.
DPO learns a Bradley-Terry reward model \citep[BT-model]{bradley1952rank} for this preference,
i.e., $p(y_1 \succ y_2 | x)$.
The Plackett-Luce reward model \citep[PL-model]{plackett1975analysis,luce1959individual}
extends BT to a multinomial case that represents a ranking over choices.
Our insight in VAs is that
general BT (and PL) reward models baked into the LLM are sufficient to perform simple tasks.
Assuming those capabilities and thus relying entirely on the accuracy of the reward models:

\def\str{\texttt{\upshape str}}

\begin{defi}[$(n,2)$-Verbalized Algorithms]
 Let $T[\tau]$ be a computational task (e.g., sorting) over objects of type $\tau$
 and let $A[\tau]$ be an algorithm that solves $T[\tau]$ with an $n$-arg boolean oracle
 $f[\tau]: \tau,\cdots\!,\tau \to \braces{\top,\bot}$.
 A \emph{Verbalized Algorithm} instantiates $A[\tau]$ with $\tau=\str$
 by implementing $f[\str]$ as a query to an LLM
 whose output is constrained to $\braces{\text{``yes''},\text{``no''}}$ which maps to $\braces{\top,\bot}$.
\end{defi}

VAs keep the high-level control flow of $A[\tau]$ intact,
while only replacing the oracle $f[\tau]$ with an LLM-based $f[\str]$.
This preserves the theoretical properties of the classical algorithm $A[\tau]$
if the model's reward model matches the ground truth.
VAs can also leverage other existing algorithmic paradigms, such as concurrent algorithms.
While the definition above focuses on the BT reward model,
VAs naturally extend to PL reward models as $(n,k)$-VAs (defined in the appendix, used in submodular maximization section).
If the reward model is inaccurate,
a non-classical robust algorithm that assumes noisy oracles
could bound the error rate of the output using known theoretical results.
A deeper investigation into the theoretical error rates is left for future work.

\textbf{Symmetrization:}
In certain tasks,
we further improve the oracles in VAs via \emph{symmetrization},
which mitigates two biases in the model:
(1) Positional bias \citep{liu2024lost}, which is the bias derived from the ordering in the input context
(e.g., always assigning a higher logit to the first choice in MCQ tasks),
and (2) sycophancy \citep{casperopen}, which is the bias toward certain answers, such as ``yes''.

For asymmetric oracles $f(x_1,x_2)$ in $(n,2)$-VAs, such as $(2,2)$-comparison oracles for sorting,
let $p_{12}$ and $p_{21}$ be next-token probabilities for answering ``yes''
with the original prompt and a prompt whose $x_1$ and $x_2$ are swapped.
Due to asymmetricity, ideally $f(x_1,x_2) = \lnot f(x_2,x_1)$ holds, but in fact $p_{12}\not=1-p_{21}$ due to the bias.
Symmetrized $(n,2)$-VAs use $(\frac{p_{12}+(1-p_{21})}{2}, \frac{(1-p_{12})+p_{21}}{2})$
as the final probability for ``yes'', ``no''.
Symmetrization for $(n,k)$-VA is discussed in the appendix.

\section{Verbalized Maximum}
\label{sec:maximum}

We first establish our claim that
\emph{even the recent sophisticated models often fail at simple algorithmic tasks}.
As an example task,
we first investigate their ability to find the maximum in
a list $X$ of $n$ randomly selected integers between 1 and 10000.
Due to the lack of principled algorithm for the LLM to follow,
a naive LLM often fails even at this simple task despite with the help of constrained decoding,
especially when the list gets longer.

We compare two approaches:
\textbf{Constraint decoding baseline} queries LLMs with a string
``What is the largest number in [comma-separated numbers in $X$]?''.
We use constraint decoding \citep{willard2023outlines} to limit the space of possible outputs to one of the elements in $X$.
\textbf{Verbalized ``Native'' maximum}
overrides Python's built-in \texttt{\textunderscore\textunderscore{}gt\textunderscore\textunderscore{}} operator
with an LLM call and applies the built-in \texttt{max} function,
which sequentially reduces the list with $O(n)$ comparisons.
Each binary comparison for $x_1,x_2\in X$ is replaced by
a yes-no style LLM query ``Is X larger than Y? X: $x_1$ Y: $x_2$'',
where the next output token is restricted to ``yes'' and ``no'' with constrained decoding.
The model output is mapped to $\top/\bot$ if the sampled output is ``yes'' / ``no'', respectively.
Finally,
\textbf{Verbalized parallel elimination tournament}
performs $\lceil\log_2 n\rceil = O(\log n)$ steps of parallel pairwise comparisons,
where each step eliminates half of the elements \citep{karp1988survey}.
As each match in the tournament eliminates 1 element, it performs $n-1$ comparisons in total.

\begin{figure}[tb]
\centering
\includegraphics[width=0.24\linewidth]{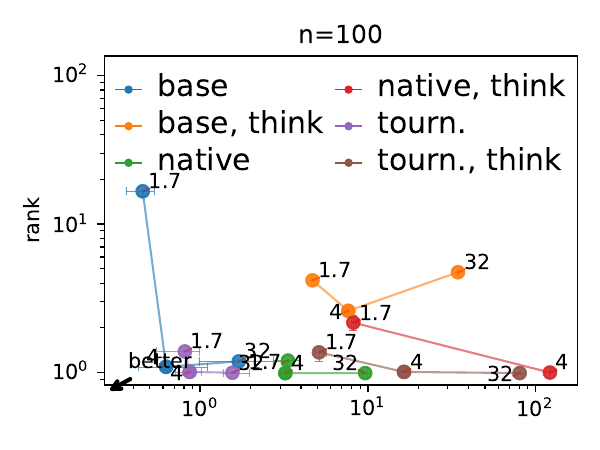}
\includegraphics[width=0.24\linewidth]{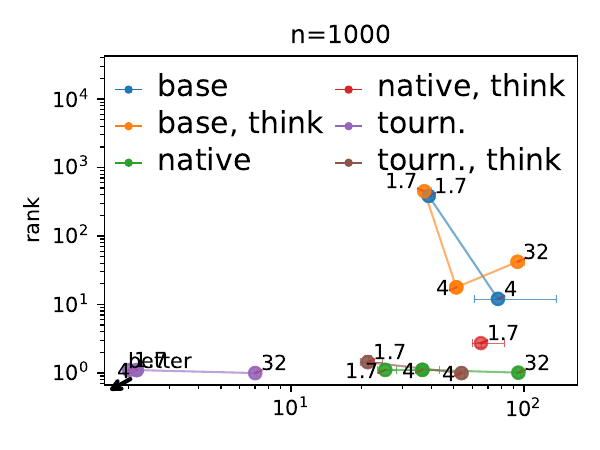}
\includegraphics[width=0.24\linewidth]{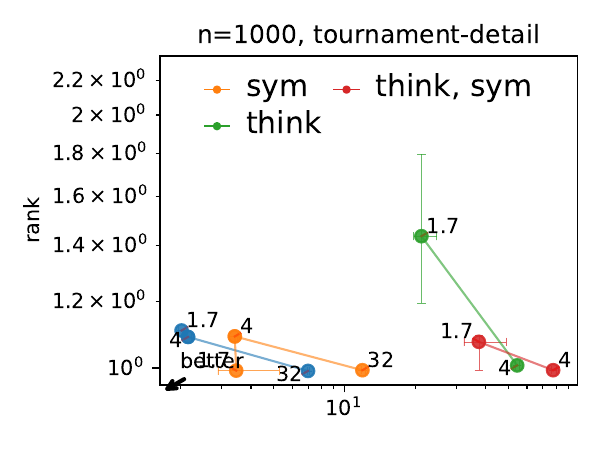}
\includegraphics[width=0.24\linewidth]{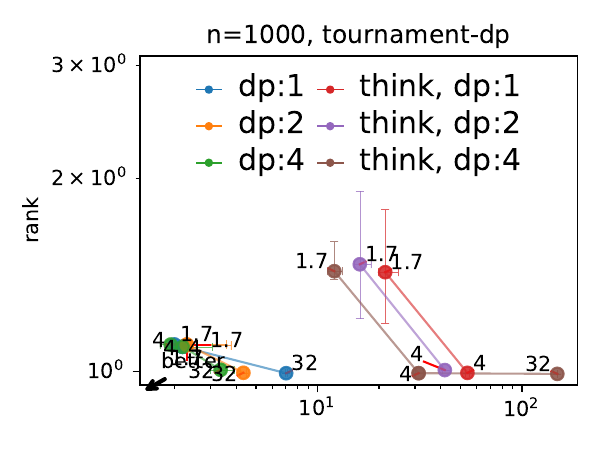}
\caption{
 The speed-accuracy Pareto front ($\log x$-axis: the runtime per query, $\log y$-axis: rank error) of maximum task
 with various model sizes (1.7B, 4B, 32B). The numbers near each point indicate the model size.
(column 1,2) \textbf{Constraint decoding baseline} vs.\ \textbf{native maximum} vs.\ \textbf{tournament maximum} with $n=100,1000$.
(column 3) Comparing various configurations of tournament on $n=1000$.
Thinking significantly increases the runtime while it can increase the error.
Symmetrization increases the runtime to a lesser degree and reduces the error.
(column 4)
The multi-GPU data parallelism (DP) on tournament on $n=1000$.
}
\label{fig:maximum}
\end{figure}

We evaluated the algorithms above across 20 randomly sampled lists for each of $n\in\braces{100, 1000}$
with various-sized Qwen3 models (1.7B, 4B, 32B) with and without thinking \citep{qwen3technicalreport}.
We measured
the \emph{rank error}, i.e., where the returned output ranked in the sorted list of $n$ numbers instead of being the first.
\refig{fig:maximum} shows the average of 5 runs.
The error bars show the minimum/maximum instead of $\text{mean}\pm\text{std}$
because the $\text{mean}-\text{std}$ can become negative and invalidate logarithmic plots.
The same error bar scheme is used throughout this paper.
In column 1 ($n=100$),
the baseline with a small 1.7B model struggles to report accurate results despite the small input size.
Thinking did not resolve the issue, and even increased the error for the larger 4B and 32B models.
Verbalized ``Native'' maximum maintains a small, often zero error.
Native+thinking worsened the error with an increased runtime, making the largest 32B model
too slow to finish 20 queries within the 1 hour time limit ($\approx$ 3 minutes per query on average).
The trend worsened in the $n=1000$ experiments (\refig{fig:maximum}, col 2).
Now thinking has little effect on the baseline and
Native+thinking timed out with the 4B model.
The only configuration that solved $n=1000$ reliably is the tournament maximum,
which was several orders of magnitude faster yet maintained near-zero errors.
Thinking helped little on the tournament either.

We further studied various configurations of tournament maximum.
\refig{fig:maximum} (mid) compares the effect of thinking and symmetrization on the runtime.
Thinking increased the runtime almost x10 fold.
In contrast, symmetrization multiplied the runtime by two, yet reduced the error in 1.7B.
Further, \refig{fig:maximum} (right) shows that
data parallelism with $\text{dp}\in\{2,4\}$ GPUs scales linearly.

\section{Verbalized Sorting}
\label{sec:sorting}

\emph{Verbalized Sorting} sorts a list of natural language strings $X$
using LLM's question answering capability as an atomic comparison oracle.
We evaluated the accuracy-speed tradeoff of several approaches.

LLM-based sorting is highly relevant to information retrieval (IR),
and the methods discussed below overlap with its existing work. 
However,
IR is not the representative of the general sorting task because of the large mismatch between their goals.
IR often targets top-$k$ retrieval and
benchmark metrics such as Normalized Discounted Cumulative Gain \citep[NDCG]{jarvelin2002cumulated}
rather than the ability to sort the entire list, including the bottom of the list, accurately.
Standard IR datasets, such as TREC / BeIR, contain binary / coarsely graded biased ranking
(majority of documents are irrelevant to the query)
where the ranking among irrelevant documents are discounted.
While some papers \citep{qin2024large} mention the time complexity of running pairwise comparisons,
evaluations predominantly focus on the accuracy and the total query cost,
and they do not analyze complexity-theoretic parallel runtime or the theoretical error rate from using noisy oracles in depth.
Among 5 recent papers \citep{qin2024large,sun2023chatgpt,sachan2022improving,ma2023zero,chao2024make},
only one table discusses the sequential runtime \citep[Table 4]{chao2024make}.

\textbf{Baseline}
encodes the list $X$ in a markdown format and asks the model to return a sorted list
using constrained-decoding
to constrain the output to be a list whose elements are guaranteed to be indices in the list.
The approach corresponds to \emph{listwise} ranking approach \citep{ma2023zero} in the RAG literature.
Due to the lack of a computationally efficient way to enforce the \emph{sorting invariances}, i.e.,
that the output
(1) is free of duplicates and (2) represents the same set of strings as the input,
we remove the duplicates and append missing elements to the end of the list.
To address the invariance violations, we also tested
\textbf{autoregressive (AR) baseline}
which decodes the sorted elements one by one,
with each step constrained to the remaining elements.
This approach is significantly slower due to the high cost of generating a new logit processor for constraining each output.

\textbf{Scoring-based approaches}
ask the model to assign a score between 1.0 and 5.0 (down to one decimal) to each sentence
using regex-constrained decoding,
then sort the sentences using the scores.
Its runtime is dominated by its $O(n)$ scoring phase.
It is a form of formalization-based approach (the list of scores represents the formal problem).
We tested two variants:
\textbf{I.i.d Scoring} scores $x_i\in X$ independently,
and \textbf{autoregressive (AR) scoring} scores all strings in a single query,
allowing the model to see the scores of other elements.
The approach corresponds to \emph{pointwise} ranking approach \citep{sun2023chatgpt,sachan2022improving} in the RAG literature.

\textbf{Verbalized sorting approaches} correspond to \emph{pairwise} ranking approaches \citep{qin2024large} in the RAG literature,
but our contribution focuses on leveraging the theoretical properties of classical algorithms,
which has been undermined in the literature.
These approaches use the same binary comparison oracle used by verbalized maximum.
\textbf{VA Bitonic}
uses the \emph{Bitonic sorting network} \citep{batcher1968sorting} from parallel sorting literature,
which performs $O(n(\log n)^2)$ comparisons in $O((\log n)^2)$ time due to parallelism.
\textbf{VA KwickSort} \citep{ailon2008kwicksort} is an \emph{approximate parallel sorting} algorithm.
It is analyzed on a more general FAS-Tournament task,
which permits transitivity violations ($a\leq b\land b\leq c \not\then a\leq c$) in LLM-based oracles,
and bounds the expected number of deordered pairs ($(\ldots, j, \ldots, i, \ldots)$ for $i<j$)
at most three times the optimal.
Operationally, kwicksort works identically to quicksort.
It runs $O(n\log n)$ comparisons in $O(\log n)$ expected parallel runtime
because the input size halves in each recursion in expectation.
To address runtime/accuracy tradeoff,
we added early-stopping to the recursion
which returns the input $X$ immediately if $|X|\leq\theta\in\{4,16\}$.

\begin{figure}[tb]
\centering
 \begin{minipage}[b]{.68\linewidth}
\includegraphics[width=.48\linewidth]{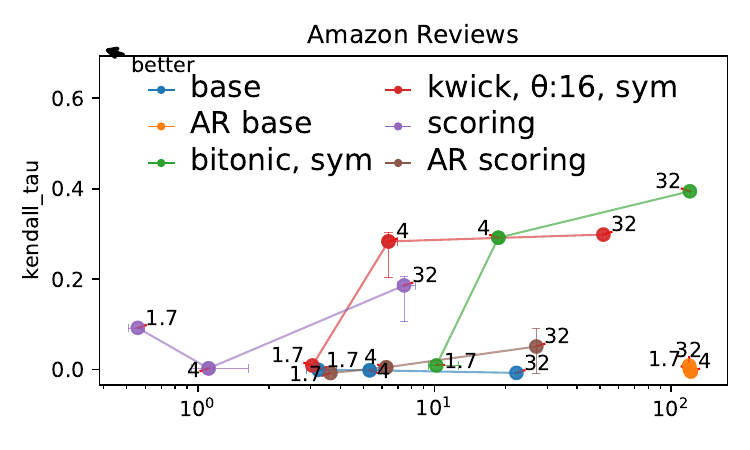}
\includegraphics[width=.48\linewidth]{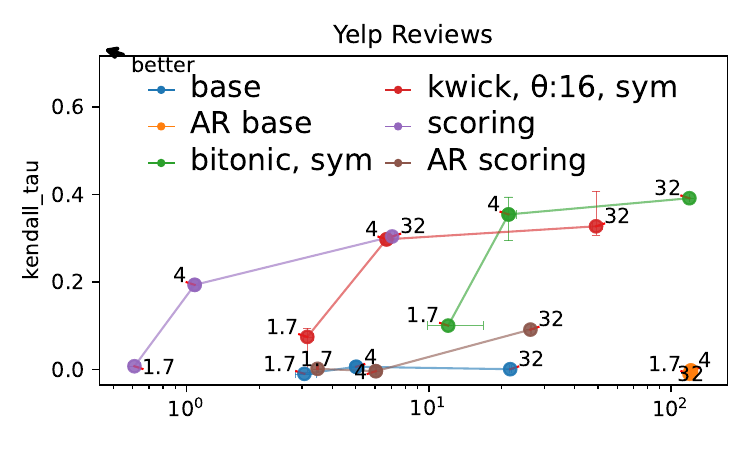}\\
\includegraphics[width=.48\linewidth]{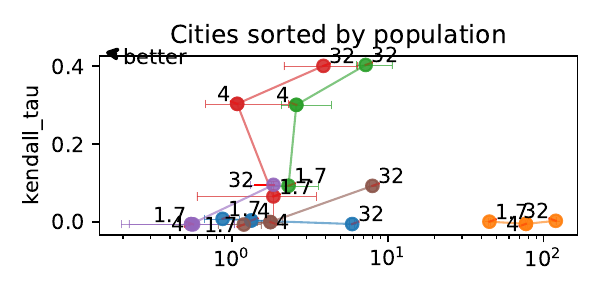}
\includegraphics[width=.48\linewidth]{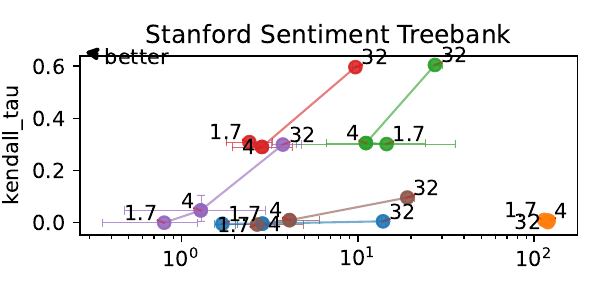}
 \end{minipage}
 \hfill
 \rule{1pt}{2in}
 \hfill
 \begin{minipage}[b]{.28\linewidth}
  \includegraphics[width=\linewidth]{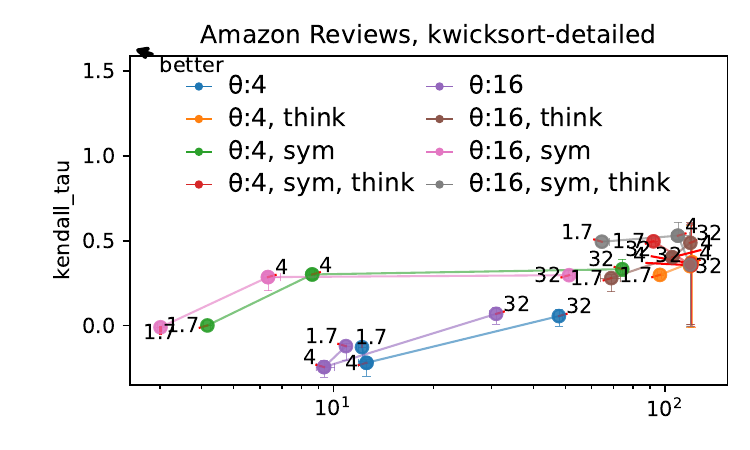}\\
  \includegraphics[width=\linewidth]{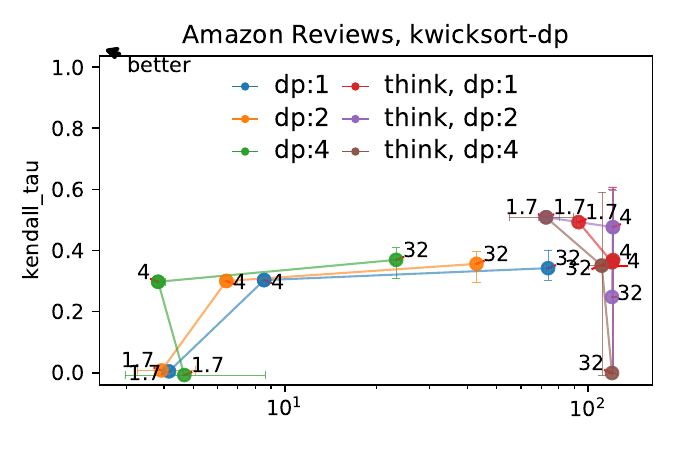}
 \end{minipage}
\caption{
 The speed-accuracy Pareto front
 ($\log x$-axis: the runtime per query, $y$-axis: Kendall-tau score) of sorting tasks.
 (left) Comparing methods.
 \textbf{kwicksort} achieved the best Pareto front.
 (right) Comparing various configurations of kwicksort on amazon reviews.
}
\label{fig:sorting}
\end{figure}

We evaluated the approaches above on the following benchmarks:
Amazon Reviews \citep{hou2024bridging},
Yelp Open Dataset Reviews \citep{yelp-open-dataset-review},
Stanford Sentiment Treebank \citep[SST]{socher-etal-2013-recursive},
and World Cities \citep{bamwor2026cities} (appendix for details).
Each dataset contains 30 sets of 100 strings,
sorted by ratings, sentiment scores, and population.
We compute the Kendall-Tau score \citep{kendall1938new} between
the output $Y$ and
the ground-truth sorted list $Y^*$,
averaged over the dataset.

\refig{fig:sorting} shows the speed-accuracy Pareto front of Qwen3 models. All VAs use symmetrization.
The baseline performed terribly in both the accuracy and the runtime.
AR baseline was even worse, with x100 slowdown from the baseline, causing timeouts (1 hour).
While the scoring obtained a reasonable accuracy with the largest 32B model,
their performance degrades as the model gets smaller.
AR scoring did not work as much and its performance is often comparable with the baseline.
VA kwicksort ($O(\log n)$) outperformed VA Bitonic ($O((\log n)^2)$) as predicted by the theory.
\emph{The strength of VA is most pronounced with the small models.}
4B VA Kwicksort outperformed 32B scoring on Amazon, Yelp, Cities, and on par in SST.

Next,
we investigated the effect of thinking, symmetrization, threshold $\theta$, and data parallelism on kwicksort.
\refig{fig:sorting} shows that
both thinking and symmetrization drastically improve the accuracy, but
thinking significantly increases the runtime
while symmetrization sometimes reduces the runtime
because of the more balanced recursions in kwicksort.
Larger $\theta\in\{4,16\}$ reduces the runtime with slight accuracy degradation.
Linear speedup from data parallelism (DP) shows the scalability of VAs.

\begin{wraptable}[7]{r}{.45\linewidth}
\vspace{-.5\baselineskip}
\begin{adjustbox}{max width=\linewidth}
 \setlength{\tabcolsep}{0.2em}
 \begin{tabular}{lccc}
dataset & 1.7B/think & 4B/think & 32B/think \\\midrule
Amazon  & 5.0/7.2    & 2.1/7.7  & 2.5/8.3   \\
Yelp    & 6.3/8.5    & 1.1/9.3  & 4.1/10.2  \\
SST     & 0.1/3.2    & 0.1/5.2  & 2.1/9.8   \\
cities  & 0.0/0.0    & 0.1/1.5  & 1.8/9.8   \\
\bottomrule
\end{tabular}
\end{adjustbox}
\end{wraptable}
Finally,
the table on the right shows that
there is no correlation between the model size and
the number of sorting invariance violations (duplicates) by the baseline.
Thinking appears to increase the violations.
This highlights the lack of soundness in free-form reasoning,
which does not occur in VAs and formalization-based approaches.

While \iid scoring is generally fast,
it is accurate only on reviews-like datasets where the ground-truth dataset also correspond to a finite range (1-5 stars).
This assumption does not apply to the population of cities which lacks a well-defined range,
demonstrating susceptibility of formalization-based approaches to the target language mismatch.
Since each list to sort has a widely different range of populations,
if all cities are small/large, the resulting score values predicted by the LLM lacks the necessary precision.
Although it may be more natural to sort the cities by directly predicting the population,
this case study is merely a representative of the data distribution
where the range of values cannot be known in advance.

Finally, we mention the effect of \emph{kv-cache reuse}.
While each VA query requires $O(t^2)$ computations for $t$ input tokens,
specialized inference engines \citep{vllm,sglang}
reuse the \emph{prefix kv-cache} over multiple queries.
This affects the resource complexity of VA queries majority of which share the prefix.
Due to space, we moved its theoretical and empirical analysis to the appendix.

\section{Verbalized Clustering}
\label{sec:clustering}

Topic clustering attempts to cluster a set of strings $S$ into a family of subsets $\mathcal{S}$ based on their similarity.
The challenge with text-based clustering is that no \emph{a priori} metric space for clustering exists.
While traditional approach toward this task performs $k$-Means or other standard clustering algorithms
on the distance matrix over the vector embeddings of the input strings,
\emph{Verbalized Clustering} instead leverage \emph{triplet comparison query} or \emph{cluster agreement query} to LLM.
Triplet comparison takes three elements $x$, $y$ and $z$ in a metric space $\Vert\cdot\Vert$ and
query if $\Vert x-z\Vert < \Vert y-z\Vert$,
i.e., we ask the LLM a question such as ``Is Z similar to X than is to Y? X: $x$, Y: $y$, Z: $z$''.
Cluster agreement query takes two elements $x$ and $y$,
then query if $x\in C \iff y \in C$,
i.e., we ask the LLM a question such as ``Does X and Y belong to the same class, category, etc.? X: $x$, Y: $y$''.
Similar to sorting, while some methods utilize LLM-based triplet comparisons for clustering
\citep{zhang2023clusterllm,viswanathan2024large},
these methods are driven by practical considerations rather than algorithmic complexity analysis.

\begin{figure}[t]
 \centering
 \begin{minipage}[c]{0.56\linewidth}
  \begin{algorithm}[H]
   \caption{Vanntec($S$, \#tree $k$, \#cluster $n$, threshold $\theta$)}
   \label{alg:vanntec}
   \begin{algorithmic}
    \For{$1\leq i<k$} ; Tree $T_i\gets \emptyset$\Comment{set of edges}
    \Flet{Rec}{parent $p$, strings $S$}
      \State \textbf{If} $|S|\leq\theta$ \textbf{then} $\forall x\in S; T_i\gets T_i\cup\{(p,x)\}$; \textbf{else}
      \State $x, y \gets $ randomly chosen from $S$
      \State $T_i\gets T_i\cup\{(p,x),(p,y)\}$
      \State $S_x \gets \braces{z \in S \mid \Vert x-z\Vert < \Vert y-z\Vert}$ 
      \State $\function{Rec}(x, S_x); \function{Rec}(y, S\setminus S_x)$
    \EndFlet
    \State $\function{Rec}$(``root'', $S$)
    \EndFor
    \State \textbf{return} EAC($n$, $\{T_1,\cdots T_k\}$)
   \end{algorithmic}
  \end{algorithm}
 \end{minipage}
 \hfill
 \begin{minipage}[c]{0.425\linewidth}
  \begin{figure}[H]
   \centering
   \includegraphics[width=\linewidth]{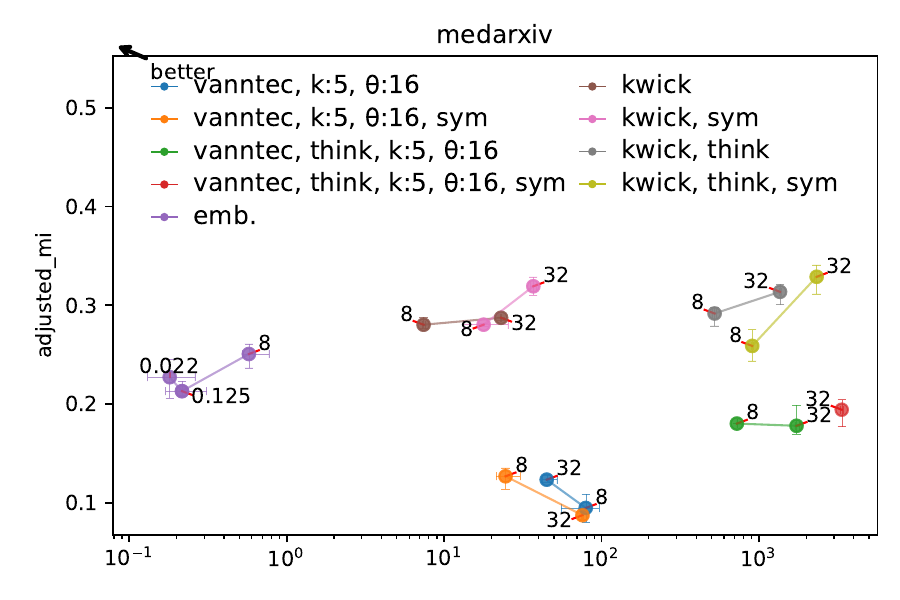}
  \end{figure}
 \end{minipage}
 \\
 \includegraphics[width=.48\linewidth]{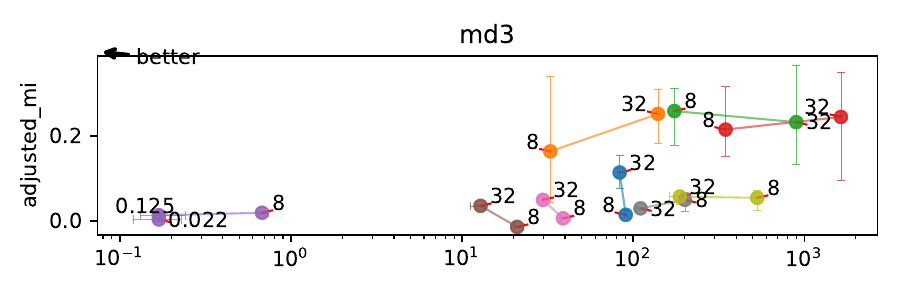}
 \includegraphics[width=.48\linewidth]{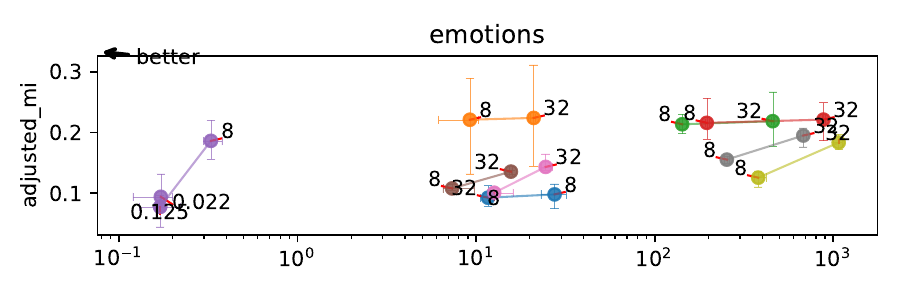}
\caption{
 The speed-accuracy Pareto front
 ($\log x$-axis: the runtime per query, $y$-axis: Adjusted Mutual Information) of clustering tasks.
}
\label{fig:clustering}
\end{figure}

We propose two verbalized algorithms.
The first is
\textbf{Verbalized Approximate Nearest Neighbor Tree Ensemble Clustering} (Vanntec, \refalgo{alg:vanntec}), which
combines $O(n\log n)$ triplet-based approximate nearest neighbor tree construction algorithm \citep{haghiri2017comparison}
which was originally developed for $k$-Approximate Nearest Neighbor Search,
and
Evidence Accumulation Clustering \citep[EAC]{fred2005eac,fred2002eac}
consensus clustering (ensemble clustering) algorithm.
It constructs multiple trees, views them as an ensemble of hierarchical clustering,
then aggregates the results to return the final clustering.
The EAC step does not involve LLM calls,
thus $O(n\log n)$ with a large constant factor for LLM calls dominates the runtime.
The parallel runtime is $O(\log n)$.
The second algorithm is \textbf{Verbalized KwickCluster} based on \citep{ailon2008kwicksort}.
In each iteration, it samples a pivot and performs a cluster agreement query
to collect elements in the same cluster.
It has the worst-case $O(n^2)$ queries and $O(n)$ parallel runtime yet practically runs much faster,
and its $O(\log n)$ parallel time improvement exists in the literature \citep{pan2015parallel}.
We compared VA clustering with 
\textbf{embedding baseline}, which
uses all-MiniLM-L6-v2/granite-embedding-125m-english/Qwen3-8B-Embedding from Sentence Transformer library
and apply K-Means \citep{macqueen1967kmeans}.

We evaluated the methods across three datasets.
\textbf{Medrxiv}
\citep{muennighoff2022mteb} contains the titles and the categories of medical papers on Medrxiv.
\textbf{MD3}
\citep{eisenstein2023md3} contains English transcripts of people from three regions with different dialects (US, India, Nigeria) playing a guessing game.
\textbf{GoEmotions}
\citep{demszky2020goemotions} contains sentences labeled with emotion tags. We sampled one tag per sentence to form clusters.

\refig{fig:clustering} shows the accuracy-runtime Pareto front
using the Adjusted Mutual Information \citep{vinh2009information,vinh2010information}.
(See appendix \refig{fig:clustering-rand} for Adjusted Rand Index \citep{hubert1985comparing} results).
Overall, VAs spent significantly longer runtime, but outperformed the baselines on accuracy.
Vanntec performed better on tasks where the clustering criteria is ambiguous
(due to emotion label sampling, or the nature of dialectal differences).
KwickCluster performed better on Medrxiv where the cluster is clear and categorical and
clustering agreement can be answered reliably.
Thinking was not cost-effective, as it only marginally helped the accuracy with x10-x100 more runtime.
On the contrary, symmetrization sometimes doubled the accuracy with a small runtime increase.
Further analysis in the appendix
\refig{fig:clustering-misc} shows the ensemble effect of Vanntec,
which demonstrates the importance of aggregating multiple trees via consensus clustering (EAC),
and the visualization of the first tree formed by Vanntec in the first list of Medrxiv dataset,
which showcases its interpretability.

\section{Verbalized Submodular Maximization}
\label{sec:vgsm}

A \emph{submodular set maximization} task is a task of
maximizing the utility $f(S)$ of a solution $S\subseteq C$ in a search space $C$,
where $f(S)$ has \emph{diminishing returns}, i.e.,
$\forall x\in C, S\subseteq S'\subseteq C; f(S'\cup \braces{x})-f(S') \leq f(S\cup \braces{x})-f(S)$.
The task has various applications including information retrieval as well as
resource allocations.
This approach, however, only works in a setting that has a well-defined submodular evaluation function $f$.
Once the set $C$ is the universe of natural language input,
and when the criteria for optimization can be only informally specified,
classical approaches toward submodular maximization are not directly applicable.
We propose \emph{Verbalized Greedy Submodular Maximization (VGSM)},
a verbalized algorithm that follows the classical greedy algorithm that guarantees $1-1/e$ optimality.
VGSM is an $(O(|C|), O(|C|))$-VA for selecting $|S|=k$ elements from $C$,
whose step $i$ selects an element from $|C|-i$ remaining elements.
It is based on chat interactions, whose step incrementally asks for a new element.
While traditionally submodular maximization requires a concrete evaluation function $f$,
VGSM specifies an implicit score function by giving a prompt to the LLM (available in the appendix).
Although it does not produce an absolute score value $f(S\cup \braces{x})$
for a newly selected element $x\in C\setminus S$,
this is not an issue because greedy algorithm
only requires a contrastive measure that encodes the selection policy.
We evaluated this approach on WiFi Access Point (AP) Optimization
and two Retrieval Augmented Generation tasks.

\paragraph{WiFi AP Optimization}

\begin{figure}[tb]
\centering
  \includegraphics[width=.4\linewidth]{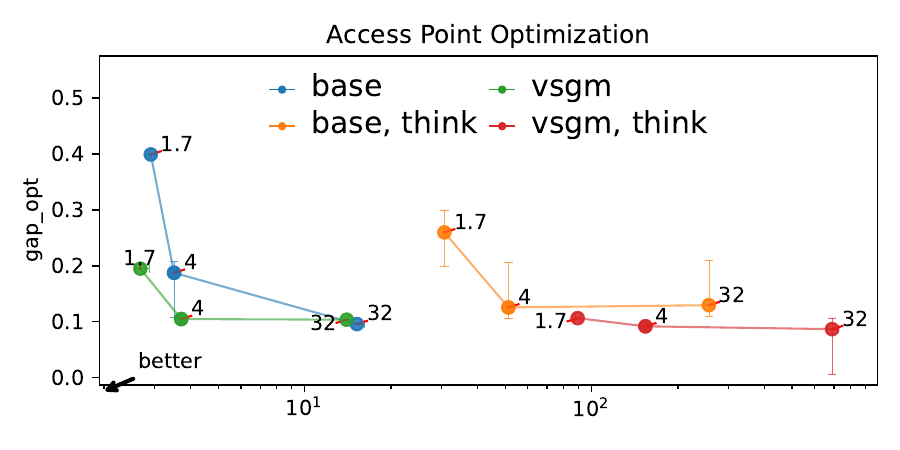}
  \includegraphics[width=.4\linewidth]{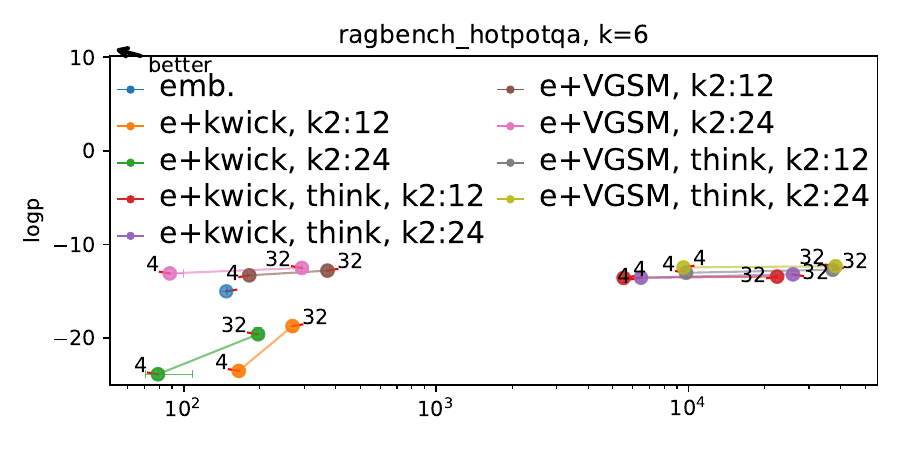}
  \\
  \includegraphics[width=.4\linewidth]{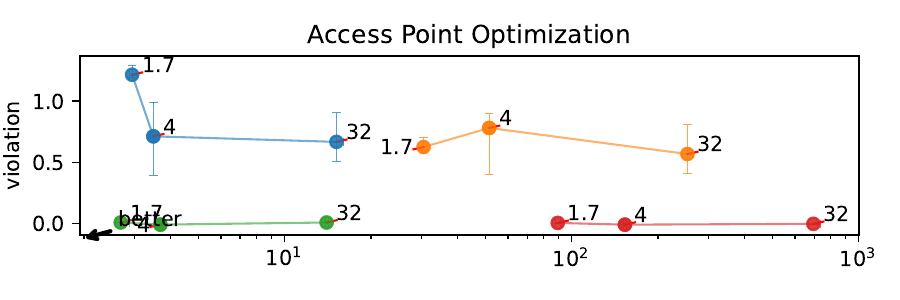}
  \includegraphics[width=.4\linewidth]{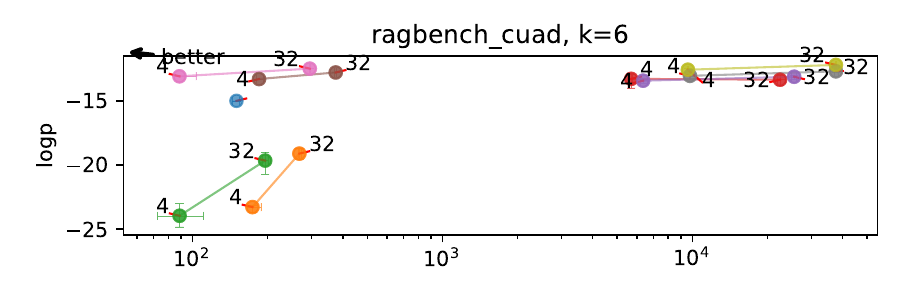}
\caption{
 The speed-accuracy Pareto front of submodular maximization tasks.
 In all subplots, $\log x$-axis is the runtime per query.
 $y$-axes are:
 (Top left) Optimality gap of WiFi tasks.
 (Bottom left) The number of AP location violations in WiFi tasks.
 (Right) \function{logp} of RAG tasks (hotpotQA, CUAD)
 retrieving $k=6$ documents.
 $k=3,12$ 
 and Jaccard distance results are available in the appendix.
 }
 \label{fig:wifi}
 \label{fig:rag}
\end{figure}

This is a classic application of submodular maximization.
Using 10 real-world floor plans obtained from HouseExpo dataset \citep{li2020houseexpo},
we created a coarse grid of height and width at least 16 cells,
then obtained the exact optimal WiFi AP locations for $k=3$ APs
using a branch-and-bound procedure
under the submodular log-distance path loss model (appendix \refsec{sec:log-distance-path-loss-model}).
Visualization of the optimal locations indicated
that the APs should be placed near the center of or the boundaries between the rooms
to provide the maximum coverage (appendix \refig{fig:wifi-floor}), matching our natural intuition.
This suggests that an LLM could return a reasonable location assignments, albeit imprecise or suboptimal, if it actually has a strong common-sense.

We compared VGSM with
a \textbf{baseline}
which uses JSON-structured decoding to return the entire assignment as a list of $(x,y)$ coordinates.
Both approaches are given the same prompt prefix
which specifies the evaluation function in python code
and an ASCII art floor map, ensuring the same amount of information is given.
We measured the runtime and the optimality gap against the optimal solution.
\refig{fig:wifi} shows that
VGSM outperformed the baselines and
obtained solutions are relatively close to the exact optimal solution.
In addition to failing to distribute APs at effective positions to maximize the coverage,
the baseline also violated the constraints specified in the prompt (\refig{fig:wifi}, right).
This includes assigning APs in the walls despite explicitly told not to,
or placing multiple APs at the same position.
These invalid APs are removed when evaluating the solution.
While the number of violations reduced with thinking,
it did not completely go away despite suffering from a significant runtime increase,
exposing the lack of soundness in LLMs.

\paragraph{Retrieval Augmented Generation (RAG)}

Given the success of VGSM in tasks with a known explicitly-defined submodular evaluation function,
we next evaluate VGSM on an evaluation function \emph{implicitly, informally} defined by the queries.
One such task is RAG,
a mainstream approach for reducing LLM hallucinations
by retrieving documents relevant to the query from a database and
appending them to the prompt so that LLMs can draw correct answers from the documents.

While existing work tends to simply rank the documents via vector embedding similarity search,
this approach fails when the task requires combining information in multiple documents to draw conclusions,
because all the top-ranked documents may be similar to each other and lack crucial information.
To address this limitation,
existing work formulated \emph{multi-hop Q\&A tasks} \citep{yang2018hotpotqa} and
observed that there are several types in each ground-truth retrieval (e.g., bridging, comparison, etc),
differing in how they chain multiple information together.
Existing work proposed various heuristic methods that additionally require fine-tuning \citep{park2025chain},
based on the observations.

On the other hand, traditional approaches to multi-hop reasoning utilize submodularity.
For example, \emph{Maximum Marginal Relevance (MMR)} rule \citep{carbonell1998use}
takes an explicitly defined evaluation function $f$ and makes it submodular.
However, in a LLM-based retriever, $f$ is only implicitly defined by the prompt,
making it difficult to apply this insight.

Greedy submodular maximization guarantees $1-1/e$ optimality under a monotone submodular function $f$.
In contrast,
due to the lack of explicit $f$, VGSM can only make a much weaker statement,
such as \emph{if $f$ is indeed submodular, then} it can guarantee $1-1/e$ optimality.
One may wonder if such a weak statement is meaningful,
but this question is \emph{logically inverted}.
The submodular assumption 
is a model simplification of the real-world task anyway.
Given a model (implicit or explicit),
\emph{it is in fact in the realm of empirical study to verify its correctness},
as was done for the log-distance model with the actual physics experiments.
Since VGSM would work if the implicit $f$ was monotone submodular and not if otherwise,
we could verify the monotone submodularity of $f$ by evaluating VGSM.

\def\logp{\function{logp}}

We evaluated two RAG datasets:
HotpotQA \citep{yang2018hotpotqa} (fullwiki, hard) and CUAD \citep{hendrycks1cuad}.
Each dataset contains a set of 4-tuples $(Q,A,S^*,D)$ where $D$ is a set of documents,
$Q$ is the question, $A$ is the ground-truth answer, and $S^*$ is the ground-truth retrieval.
We assume a generative model
$p(A | D, Q) = \sum_S p_{\phi}(A | S, Q) p_{\psi}(S | Q, D)$,
where $S$ is the documents selected by the retrieval model $\psi$,
and $p_{\phi}(A | S, Q)$ represents the likelihood of the answer by an answering model $\phi$.
Given $\psi$ and $\phi$,
we retrieve $k=3,6,12$ documents with $\psi$ and
evaluate the performance by
$\logp=\log p_{\phi}(A | S, Q) = \sum_t \log p_{\phi}(A_{t} | A_{1:t-1}, S, Q)$,
i.e., the sum of logits over the tokens in $A$.
This logits are obtained by feeding $\phi$ a prompt concatenating $Q, S, A$ in order.
If $S\approx S^*$, the likelihood should be higher.
We also measured the Jaccard distance 
between $S$ and $S^*$.

We compared VGSM with 2 baselines.
\textbf{Embedding} baseline uses \lsota Qwen3-Embedding-8B for top-$k$ documents retrieval with cosine distance.
\textbf{Verbalized KwickSelect} uses a symmetrized parallel KwickSelect algorithm,
a top-$k$ selection variant of kwicksort
(when the right branch $R$ has $|R|>k$ elements, recurse into it;
 otherwise, recurse into the left branch $L$ and select top-$|L|-k-1$ elements).
We evaluate it in order to see the impact of ignoring the submodularity.
In VGSM and KwickSelect, we tested $\psi\in\{\text{Qwen3-4B},\text{Qwen3-32B}\}$ with $\phi=\text{Qwen3-32B}$.
To reduce the runtime of the experiment,
VGSM and Verbalized KwickSelect pre-selects $2k$ or $4k$ documents using the embedding baseline
before running the respective algorithm.

\refig{fig:rag} shows that
$k$ documents retrieved by VGSM from $2k$ or $4k$ documents have better $\logp$
than those simply selected by the embedding.
KwickSelect has worse $\logp$ with similar runtime,
thus the implicit objective encoded by the query was indeed approximately submodular.
Thinking was not cost-effective;
VGSM already took care of the sequential algorithmic thoughts,
thus there is significantly less need for thinking in each document selection step.

\section{Conclusion, Discussion, Future Work}

We demonstrated the feasibility of \emph{Verbalized Algorithms} (VAs),
a general disciplined paradigm for performing algorithmic reasoning with LLMs.
It leverages the strong prior and the established theoretical results of classical discrete algorithms
to significantly improve the accuracy even with a small model without costly training,
and improve the runtime of such reasoning.
While this paper explored only a limited number of examples due to space,
our experiments demonstrated that
overall,
spurious free-form self-thought via think-tags could harm the accuracy,
and theoretically grounded algorithmic reasoning targeted to specific tasks tends to be superior.
We would like to remind the community that
algorithms and computers were born out of humanity's particular weakness in long sequential reasoning,
and that algorithmic analysis is a tool developed to understand it.
These tools are still useful even on opaque LLM-based reasoning, and throwing them out is not an option.

Future work includes extending VA to more complex, potentially combinatorial algorithms,
including Pareto front computation for diverse / non-dominated document retrieval,
DPLL/CDCL, local search, best-first search for combinatorial reasoning / optimization / planning.

\fontsize{9.5pt}{10.5pt}
\selectfont

\clearpage
\appendix
\section*{Appendix}

\section{$(n,k)$-Verbalized Algorithms}

\def\str{\texttt{\upshape str}}

A verbalized algorithm that utilizes the PL reward model of an LLM by
using it as a $k$-categorical oracle is called a \emph{$(n,k)$-Verbalized Algorithm}.
For brevity, we may refer to $(n,k)$-VAs as just $k$-VAs.

\begin{defi}[$(n,k)$-Verbalized Algorithms]
 Let $A[\tau]$ be an algorithm that solves $T[\tau]$ with an $n$-arg boolean oracle
 $f[\tau]: \tau,\cdots\!,\tau \to 1..k$.
 A \emph{$(n,k)$-Verbalized Algorithm} instantiates $f[\str]$ as a multiple-choice-question (MCQ) query to an LLM.
\end{defi}

\paragraph{Symmetrization for $(n,k)$-VAs}

Let $\sigma$ be a permutation of $n$ inputs.
An $n$-ary function $f$ is \emph{permutation-invariant} if $\forall \sigma; f(\sigma(\vx)) = f(\vx)$, and
\emph{permutation-equivariant} if $\forall \sigma; f(\sigma(\vx)) = \sigma(f(\vx))$.
For permutation-invariant oracles in $(n,k)$-VAs,
we query them with $S$ shuffled inputs (hyperparameter $S\leq n!$)
presenting the arguments in a different order in the context,
then average the output probability over the orderings.
For permutation-equivariant oracles, the probability must be added toward the original element before the shuffling.

\let\_\textunderscore

\section{Sorting Dataset and Prompt Details}
\label{sec:sorting-details}

This section summarizes the dataset used for evaluating the sorting task.
All verbalized sorting algorithms use the same query for the same dataset.
Let \texttt{comparison\_prompt} be a comparison query prompt, specific to each dataset, listed in the following sections.
When comparing strings \texttt{X} and \texttt{Y},
we query the LLM the following prompt, using the f-string notation in Python programming language:

\begin{minted}[fontsize=\footnotesize]{python}
 verbalized_prompt = f"{comparison_prompt}\nX:{X}\nY:{Y}"
\end{minted}

To guarantee that all baseline approaches
(constraint-decoding baseline,
autoregressive constraint-decoding baseline,
I.i.d. scoring, autoregressive scoring) receive the same information
as the verbalized sorting, we construct the prompts for these baselines
from the same comparison query prompt used for verbalized sorting.
Let \texttt{strings} be a list of strings to sort.
For constraint-decoding baseline, we query the LLM as follows,
with outputs constrained to string representations of numbers from 0 to $|\texttt{strings}|$:

\begin{minted}[fontsize=\footnotesize]{python}
 constraint_decoding_basline_prompt = (
      f"Sort the following list in the increasing order of quality " +
      f"that is compared in the following comparison criteria: '{comparison_prompt}'. " +
      f"That is, for a pair X and Y, if the answer to the criteria is 'yes', " +
      f"X must appear after Y in the output list. " +
      f"The elements in the list are numbered. " +
      f"Answer the order using a comma-separated list of the numbers.\n" +
      "\n".join([f"{i}: {s}" for i, s in enumerate(strings)])
 )
\end{minted}

For scoring-based baseline, we query the LLM with output constrained to
a regular expression \verb`"[1-4]\.[0-9]"` as follows:

\begin{minted}[fontsize=\footnotesize]{python}
 scoring_prompt = (
   "I want to sort several strings. " +
   "Given two strings X and Y, " +
   "my comparison criteria is as follows: '" + comparison_prompt + "'. " +
   "Assign a score to X below so that for any pair of strings X and Y, " +
   "if the answer to the comparison criteria above is yes, " +
   "the score of X is greater than the score of Y. " +
   "Each score must be between 1 and 5 with one decimal place, " +
   "where the score 5 means that X will most certainly satisfy the comparison against other strings, " +
   "and the score 1 means the opposite. " +
   "For example, if we compare the height of mountains, score 5 will be assigned to the tallest mountains, " +
   "and if we compare the quality of restaurants, score 1 implies that the food or the service is very bad. " +
   "\n" +
   "X: " + X + "\n"
 )
\end{minted}

\subsection{Amazon reviews sorted by star ratings}

We downloaded the dataset from the Hugging Face repository
\texttt{McAuley-Lab/Amazon-Reviews-2023} \citep{hou2024bridging}
and extracted the \texttt{Video\_Games} category.
We discarded reviews with empty text or title or reviews with no helpful vote.
We deduplicated reviews based on identical text.
We then computed per-product rating statistics and retained only
products with more than 100 total reviews and at least 20 reviews in
each star rating class (1 to 5 stars).
After filtering products,
we reduced class imbalance within reviews per product
by limiting each rating bucket to at most 20 reviews.
We then selected 30 products from the filtered set and
truncate each review text to 1000 characters to standardize input length.

\textbf{Comparison query}
\texttt{comparison\_prompt = } ``You will be given two product reviews, X and Y, written by various customers about the same product.
Is X more favorable to the product than Y?''

\subsection{Yelp reviews sorted by star ratings}

We constructed the dataset from the Hugging Face dataset
\texttt{yashraizad/yelp-open-dataset-reviews}.
We filtered businesses with at least 100 reviews in total
and at least 20 reviews for each star rating (1 through 5),
then balanced the number of reviews per star (20 reviews per star per product).
We selected the first 30 businesses to form the final dataset.

\textbf{Comparison query}
\texttt{comparison\_prompt = } ``You will be given two Yelp reviews, X and Y, of the same restaurant written by various customers.
Is X more favorable to the restaurant than Y?''

\subsection{World cities sorted by population}

We constructed the dataset from the Hugging Face dataset
\texttt{bamwor/world-cities}.
We removed duplicate city names.
We retained only timezones that
contains more than 30 cities, then randomly sampled 30 such timezones. For
each selected timezone, we sampled up to 100 cities and sorted them by
population.

\textbf{Comparison query}
\texttt{comparison\_prompt = } ``You will be given two city names, X and Y, in the same timezone. Is X's population larger than that of Y?''

\subsection{Stanford Sentiment Treebank (SST) sorted by sentiment}

We constructed the dataset from the Hugging Face dataset
\texttt{stanfordnlp/sst}.
We removed duplicate sentences.
Then we formed 20 groups by randomly sampling sentences without replacement.
For each group, we sampled up to 100 sentence-label pairs and sorted them by sentiment label.

\textbf{Comparison query}
\texttt{comparison\_prompt = } ``You will be given two sentences, X and Y.
Is X more positive than Y?''

\section{Clustering Dataset and Prompt Details}
\label{sec:clustering-details}

The prompt for Triplet Approximate Nearest Neighbor Tree for \texttt{X}, \texttt{Y}, \texttt{Z} is
constructed from \texttt{triplet\_comparison\_prompt} as follows:

\begin{minted}[fontsize=\footnotesize]{python}
 triplet_anns_prompt = f"{triplet_comparison_prompt}\nZ:{Z}\nX:{X}\nY:{Y}"
\end{minted}

The prompt for kwickcluster algorithm given \texttt{X} and \texttt{Y}
is constructed from \texttt{class\_agreement\_prompt} as follows:

\begin{minted}[fontsize=\footnotesize]{python}
 kwickcluster_prompt = f"{class_agreement_prompt}\nX:{X}\nY:{Y}"
\end{minted}

\subsection{MTEB/LLM-Eval Medrxiv clustering dataset}

This dataset is readily available from Huggingface
(\texttt{mteb/llm-eval-medrxiv\_clustering\_s2s\_v2})
and is used without additional processing.
Each dataset contains 5 sets to perform clustering, and
each set has 200 strings.

\texttt{triplet\_comparison\_prompt = }
You will be given three paper titles X, Y, and Z.
Is the topic in Z closer to the topic in X than to the topic in Y?

\texttt{class\_agreement\_prompt = }
You will be given two paper titles X and Y posted on medrxiv.
Is the subject area (e.g., Orthopedics) of X same as the subject area of Y?

\subsection{MD3}

We constructed the dataset from the Hugging Face dataset \texttt{WillHeld/MD3}.
We treated the three dialects (\texttt{en\_in}, \texttt{en\_ng}, \texttt{en\_us}) as class labels.
We generated 5 sets containing 99 samples each, randomly selecting 33 samples per class.
We truncated each transcript to at most 500 characters.

\texttt{triplet\_comparison\_prompt = }
You will see three dialogue transcripts, X, Y, and Z, of people playing a guessing game.
Does the dialect in Z sound closer to the dialect in X than to the dialect in Y?

\texttt{class\_agreement\_prompt = }
You will see two dialogue transcripts, X and Y, of people playing a guessing game.
Does X use the same dialect as Y does?

\subsection{GoEmotions}

We constructed the dataset from the Hugging Face dataset \texttt{google-research-datasets/go\_emotions}.
We assigned each text to a single label by randomly selecting one of its annotated emotion labels,
then grouped texts into label-specific buckets.
We filtered out emotion labels that do not contain enough examples to support balanced sampling.
We then constructed 5 sets, each containing 100 samples drawn from 5 distinct emotion labels per set.
For each selected label, we sampled 20 texts.

\texttt{triplet\_comparison\_prompt = }
You will see three sentences, X, Y, and Z.
Does the emotion in Z sound closer to the emotion in X than to the emotion in Y?

\texttt{class\_agreement\_prompt = }
You will see two sentences, X and Y.
Does X express the same emotion as Y does?

\section{Verbalized Greedy Submodular Maximization Prompting Detail}
\label{sec:submodular-details}

VSGM performs a multi-turn conversation with the LLM.
Given a user-defined task-specific \emph{task prompt} which explains the task,
the first turn ($i=0$) queries the LLM an \emph{initialization prompt}.
Let $C$ be the universe of the set to retrieve $k$ items from,
and \verb`task_prompt` and \verb`init_prompt` be a task / initialization prompt, respectively.
Then:

\begin{minted}[fontsize=\footnotesize]{python}
 init_prompt = (
     task_prompt
     + "\n".join([ f"X{j}: {elem}" for j, elem in enumerate(C)])
     + f"\nStep {i}: Select Y{i}.\n"
     + f"Your answer must be one of: "
     + ", ".join([ "X"+str(j) for j in range(len(C))])
 )
\end{minted}

The output is constrained to \texttt{"X0"}, \texttt{"X1"}, $\cdots$\texttt{"X\{$|C|-1$\}"}.
In the later step $i>0$ of the multi-turn prompt,
Let $U$ and $S$ be the lists of
the remaining unselected elements and the selected elements that represent a partial solution.
We query the following:

\begin{minted}[fontsize=\footnotesize]{python}
 prompt = (
     "OK, the current set is "
     + ", ".join([ "X"+str(j) for j, _ in U])
     + f".\nStep {i}: select Y{i}.\n"
     + f"Your answer must be one of: "
     + ", ".join([ "X"+str(j) for j, _ in S])
 )
\end{minted}

\subsection{Prompts used in the WiFi Access Point Optimization task}
\label{sec:wifi-details}

Both algorithms (baseline, VGSM) 
use the same prompt prefix shown below:

\begin{minted}[fontsize=\footnotesize]{python}
def common_prompt(k):
    return (
        f"Your task is to put {k} wifi access points (APs) on this floor to maximize the coverage " +
        "under Log-distance path loss model, i.e., " +
        "assume the signal strength of a particular position is " +
        "the maximum signal strengths across the APs, and " +
        "the signal diminishes logarithmically to the distance between the position and the AP and " +
        "linearly to the number of walls on the line between the position and the AP. " +
        "The final score (coverage) is the sum of signal strengths over the entire floor. " +
        "This coverage function is submodular. " +
        "More concretely, it is computed by the following python code: \n" +
        """
def signal_strength(
    grid: np.ndarray,
    ap: tuple[int, int],
) -> np.ndarray:
    path_loss_exp: float = 2.0
    wall_penalty: float = 5.0
    H, W = grid.shape
    signal = np.full((H, W), -np.inf)

    for i in range(H):
        for j in range(W):
            if grid[i, j] == 0:
                continue

            d = np.sqrt((i - ap[0])**2 + (j - ap[1])**2) + 1e-6
            walls = count_walls(grid, ap, (i, j))

            signal[i, j] = -10 * path_loss_exp * np.log10(d) - wall_penalty * walls

    return signal

def coverage_function(grid: np.ndarray, aps: list[tuple[int, int]], threshold: float = -70.0) -> float:
    H, W = grid.shape
    best_signal = np.full((H, W), -np.inf)

    for ap in aps:
        sig = signal_strength(grid, ap)
        best_signal = np.maximum(best_signal, sig)

    utility = np.maximum(best_signal, threshold)
    return float(utility.sum())
        """ +
        "\nLook at the floor map shown below. " +
        "The letter '#' indicates a wall, and the letter '.' indicates an empty space. " +
        "Assume that the top-left position is the origin. "
    )
\end{minted}

Then each algorithm receives the following prompts.
Let \verb`floor_text` be an ascii art representation of the floor map.
VSGM uses the following prompt:

\begin{minted}[fontsize=\footnotesize]{python}
 task_prompt = (
     common_prompt(k) +
     f"Select {k} locations Y0..Y{k-1} " +
     f"from a list of locations X0..X{len(C)-1} one by one. " +
     "Answer the location of each AP as a x-y coordinate on the floor map. " +
     "Note that the top-left position is the origin. " +
     "\nHere is the floor map: \n" +
     floor_text +
     "\nHere are the names of available postions:\n")
\end{minted}

The baseline uses the following prompt:

\begin{minted}{python}
 prompt = (
     common_prompt(k) +
     "Answer the locations of APs as a list of x-y coordinates on the floor map " +
     "in the 'coordinates' value of a json dictionary. " +
     "Do not select the same location twice. " +
     "Do not select the locations in the walls. " +
     "\nHere is the floor map: \n" +
     floor_text)
\end{minted}

Its output is constrained to the following Pydantic-based JSON schema:

\begin{minted}{python}
 class Coordinate(BaseModel):
     x: Annotated[int, Field(ge=0,lt=W)]
     y: Annotated[int, Field(ge=0,lt=H)]
 class WifiOutput(BaseModel):
     coordinates: Annotated[list[Coordinate],
                            Field(min_length=k,
                                  max_length=k)]
\end{minted}

\subsection{Prompts in the RAG task}
\label{sec:rag-details}

The task prompt for VSGM in the RAG task is as follows:

\begin{minted}[fontsize=\footnotesize]{python}
 task_prompt = (
     f"Your task is to select {k} documents Y0..Y{k-1} "
     + f"from a list of documents X0..X{len(D)-1} "
     + "that helps answer a question Q. "
     + f"Select them in {k} steps. "
     + "In each step i, select Yi "
     + "based on what information is new on top of the documents that are already selected and "
     + "how the new information helps answer Q. "
     + "\n"
     + f"Q: {Q}\n"
 )
\end{minted}

Kwickselect performs Verbalized kwickselect algorithm for top-$k$ selection.
It uses the comparison oracle which performs a following query for each pair $(X,Y)$:

\begin{minted}[fontsize=\footnotesize]{python}
  comparison_criteria = (
      "We are collecting a set of documents that helps answer a question Q. "
      "Is information X more helpful than information Y to answer Q?"
      "\n"
      f"Q: {Q}\n"
      f"X:\n{X}\n"
      f"Y:\n{Y}\n"
  )
\end{minted}

\subsection{Ground Truth Cost Model of Wifi Task}
\label{sec:log-distance-path-loss-model}

Let $F$ be points on a $H\times W$ discrete grid representing the floor map.
Given $k$ APs at locations $\vy_1,\cdots\vy_k\in F$,
let the signal received from $i$-th AP at location $\vx\in F$ be
$S(\vy_i, \vx)=\alpha + \beta\log ||\vy_i - \vx||_2 + \gamma W(\vx, \vy_i)\in \R$,
where $W(\vx, \vy_i)$ is the number of wall grids on the straight line path between $\vx$ and $\vy_i$.
$\alpha,\beta,\gamma$ are constants.
The optimization objective is $\sum_{\vx\in F} \max_i S(\vy_i, \vx)$, which is submodular.

\section{Hardware Details}

Maximization and sorting experiments are performed
on NVIDIA Tesla A100-80GB GPUs,
while clustering and submodular maximization experiments are performed on
NVIDIA Tesla H100 GPUs.

\section{Effect of Prefix Caching on Sorting Task}

\begin{figure}[tb]
\centering
 \begin{minipage}{.8\linewidth}
  \includegraphics[width=\linewidth]{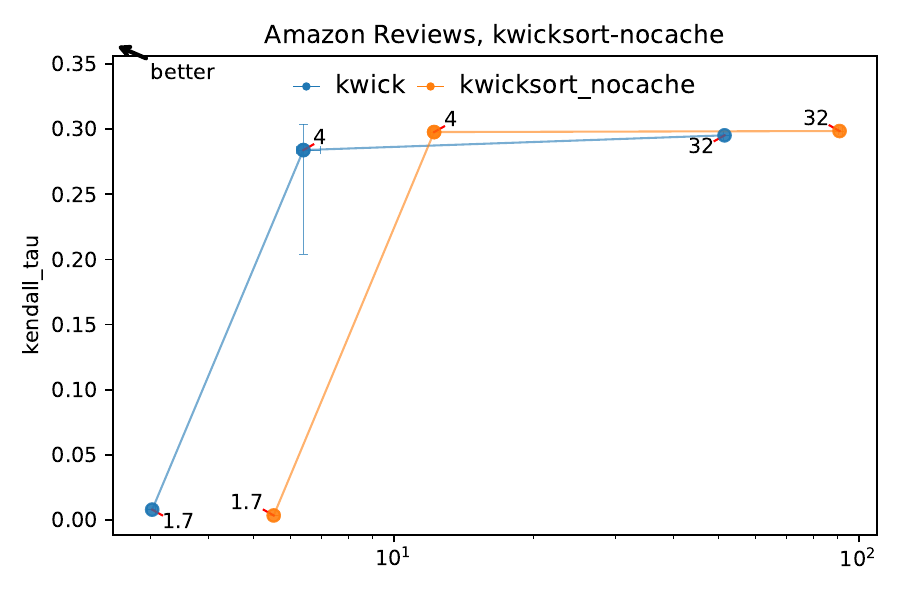}
 \end{minipage}
\caption{
 The speed-accuracy pareto front
 ($\log x$-axis: the runtime per query, $y$-axis: kendall-$\tau$ score) of sorting tasks
 over various Qwen3 model sizes (1.7B, 4B, 32B),
 comparing the effect of kv-cache reuse in KwickSort (with symmetrization, $\theta=16$).
 Disabling kv-cache reuse causes a significant runtime increase.
}
\label{fig:sort-kvcache}
\end{figure}

A comparison query in VA sorting consists of a prompt, $x_1$, and $x_2$.
Assume all strings being compared has a fixed length $L=|x_1|=|x_2|$, and that the length of the fixed prefix is $L_p$.
In the exhaustive pairwise comparison among $n$ strings ($O(n^2)$ comparisons, equivalent to insertion sort etc.),
a naive inference engine requires $O((L_p+2L)^2)$ runtime per each yes/no query,
thus $O(n^2(L_p+2L)^2)$ runtime in total, without parallelism.

We analyze the runtime reduction due to kv-caches.
Assume that an input query consists of $N+M$ tokens (prefix and suffix),
and the kv-cache of $N$ prefix tokens is already available.
Each $N+t$-th new input token attends to $N+(t-1)$ existing tokens,
which makes $N+M+1$-th token require $\sum_{t=1}^{M} (N + (t-1)) = O(NM+M^2)$ runtime.
In other words,
the runtime of each comparison query reduces to $O(2L_pL + 4L^2)$,
totalling $O(L_p^2 + n^2 (2L_pL + 4L^2))$ runtime.
Moreover, since $x_1$ is repeated $n$ times over $n^2$ comparisons,
it further reduces to
$O(L_p^2 + n (L_pL + L^2) + n^2 ((L_p+L)L + L^2))=O(L_p^2 + n (L_pL + L^2) + n^2 (L_pL + 2L^2))$.

Although we assumed a simplified, exhaustive ($O(n^2)$) comparison scenario,
this speedup is significant in practice.
\refig{fig:sort-kvcache} demonstrates the empirical speedup due to kv-cache.
RadixAttention in SGLang is particularly helpful for VAs whose queries share large prefixes.

\section{Additional Results}

The rest of the pages contain additional tables and figures.

\begin{figure}[p]
\centering
 \begin{minipage}{.8\linewidth}
  \includegraphics[width=\linewidth]{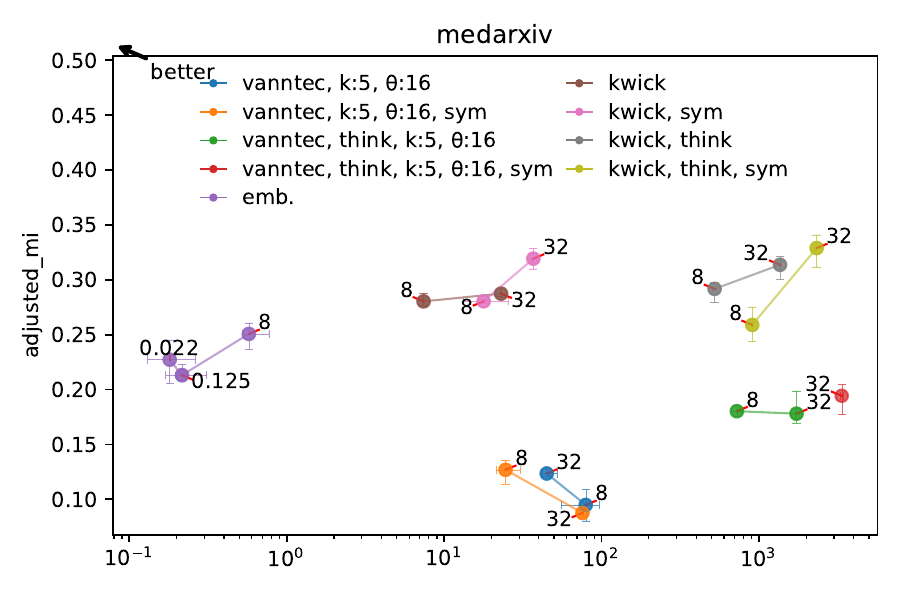}
  \includegraphics[width=\linewidth]{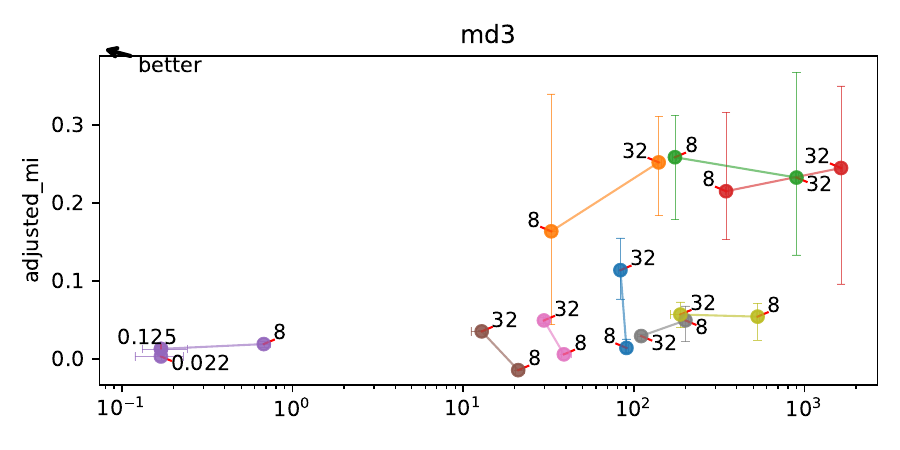}
  \includegraphics[width=\linewidth]{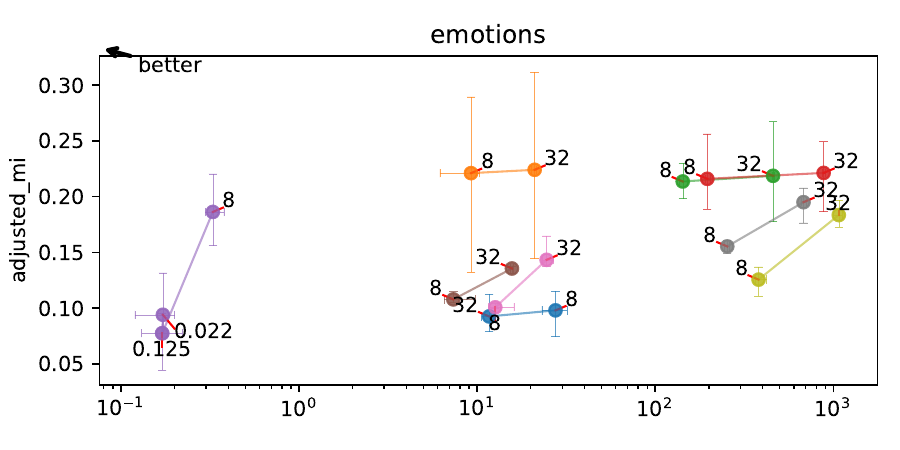}
 \end{minipage}
\caption{
 \emph{Enlarged version of \refig{fig:clustering}.}
 The speed-accuracy pareto front
 ($\log x$-axis: the runtime per query, $y$-axis: Adjusted Mutual Information) of clustering tasks
 (GoEmotions, MD3, Medrxiv) over various Qwen3 model sizes (1.7B, 4B, 32B).
 \textbf{Embedding baseline} is cheap but consistently outperformed.
 In Medrxiv, the best pareto front was obtained by
 \textbf{Kwickcluster} because the clusters is inherently categorical and determining the cluster agreement is easy.
 In GoEmotions/MD3,
 \textbf{VANNS} achieved the best pareto front
 because the clusters are inherently ambiguous/continuous and hard to quantify (dialects in MD3),
 or because the labels are sampled from the multi-label classification (GoEmotions).
}
\label{fig:clustering-mi}
\end{figure}

\begin{figure}[p]
\centering
 \begin{minipage}{.8\linewidth}
  \includegraphics[width=\linewidth]{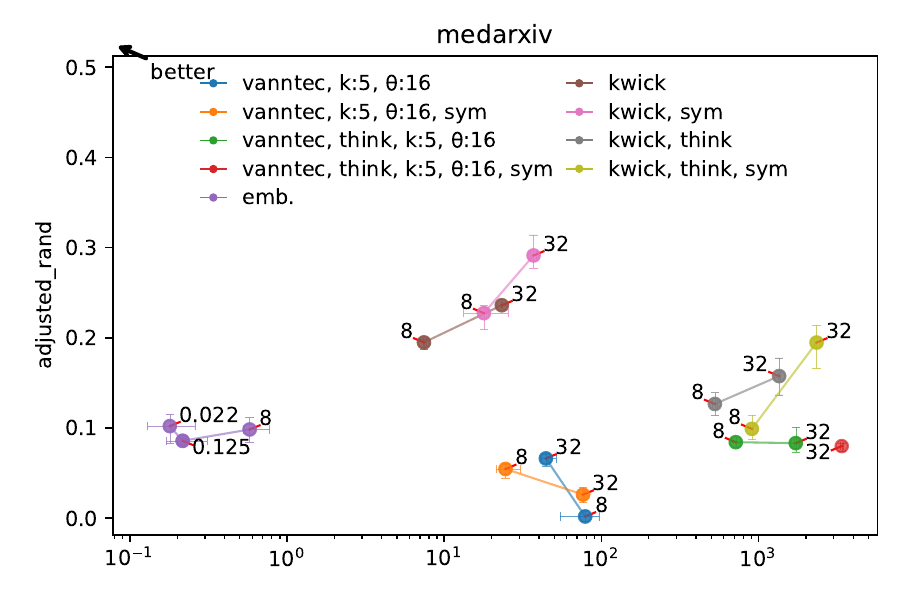}
  \includegraphics[width=\linewidth]{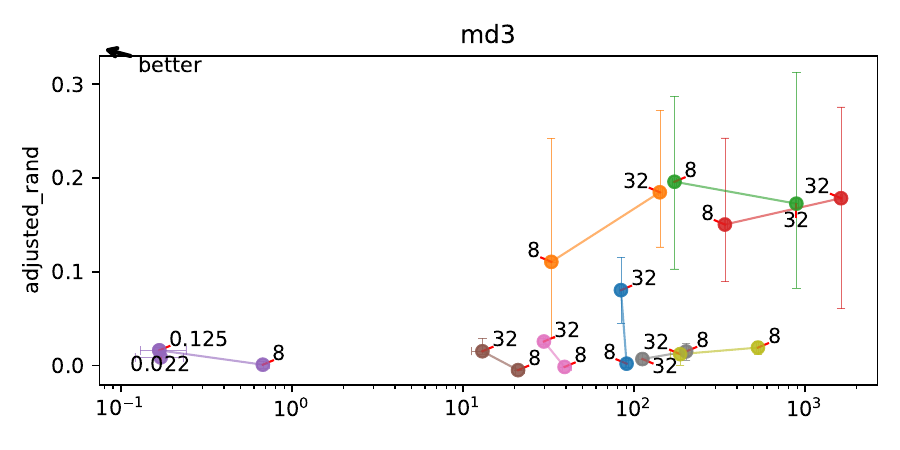}
  \includegraphics[width=\linewidth]{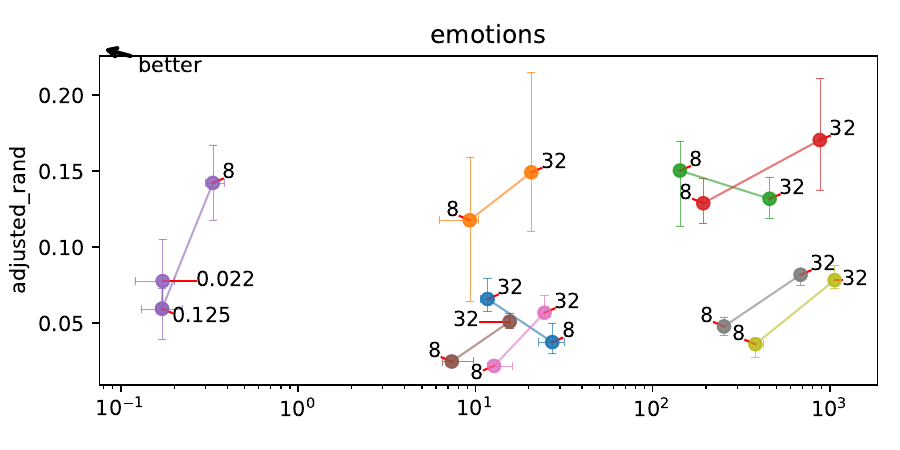}
 \end{minipage}
\caption{
 The speed-accuracy pareto front
 ($\log x$-axis: the runtime per query, $y$-axis: Adjusted Rand Index) of clustering tasks
 (GoEmotions, MD3, Medrxiv) over various Qwen3 model sizes (1.7B, 4B, 32B).
 \textbf{Embedding baseline} is cheap but consistently outperformed.
 In Medrxiv, the best pareto front was obtained by
 \textbf{Kwickcluster} because the clusters is inherently categorical and determining the cluster agreement is easy.
 In GoEmotions/MD3,
 \textbf{VANNS} achieved the best pareto front
 because the clusters are inherently ambiguous/continuous and hard to quantify (dialects in MD3),
 or because the labels are sampled from the multi-label classification (GoEmotions).
}
\label{fig:clustering-rand}
\end{figure}

\begin{figure}[p]
 \centering
 \includegraphics[width=0.9\linewidth]{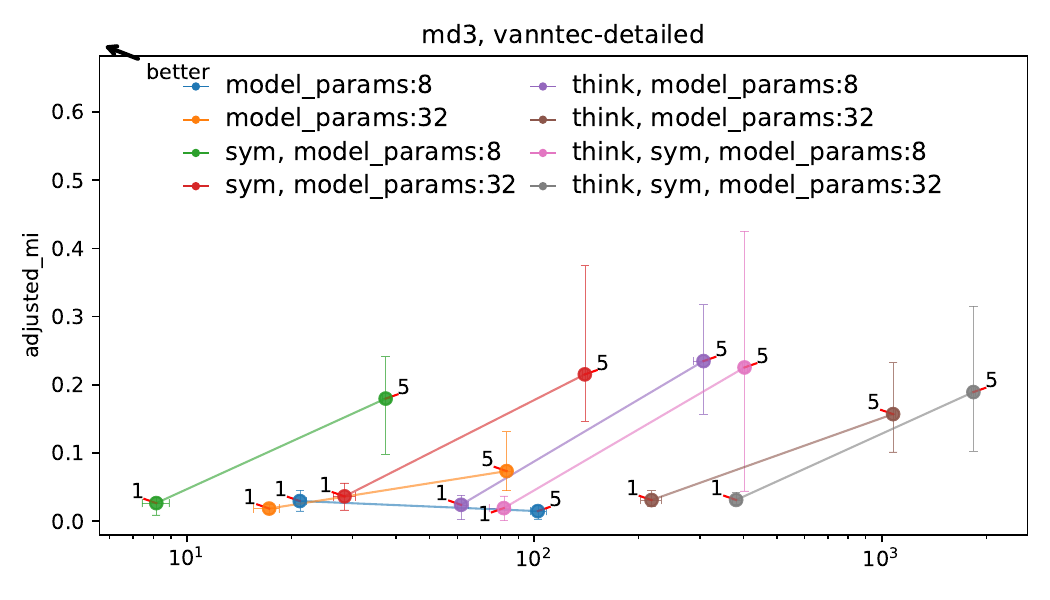}\\
 \includegraphics[width=\linewidth]{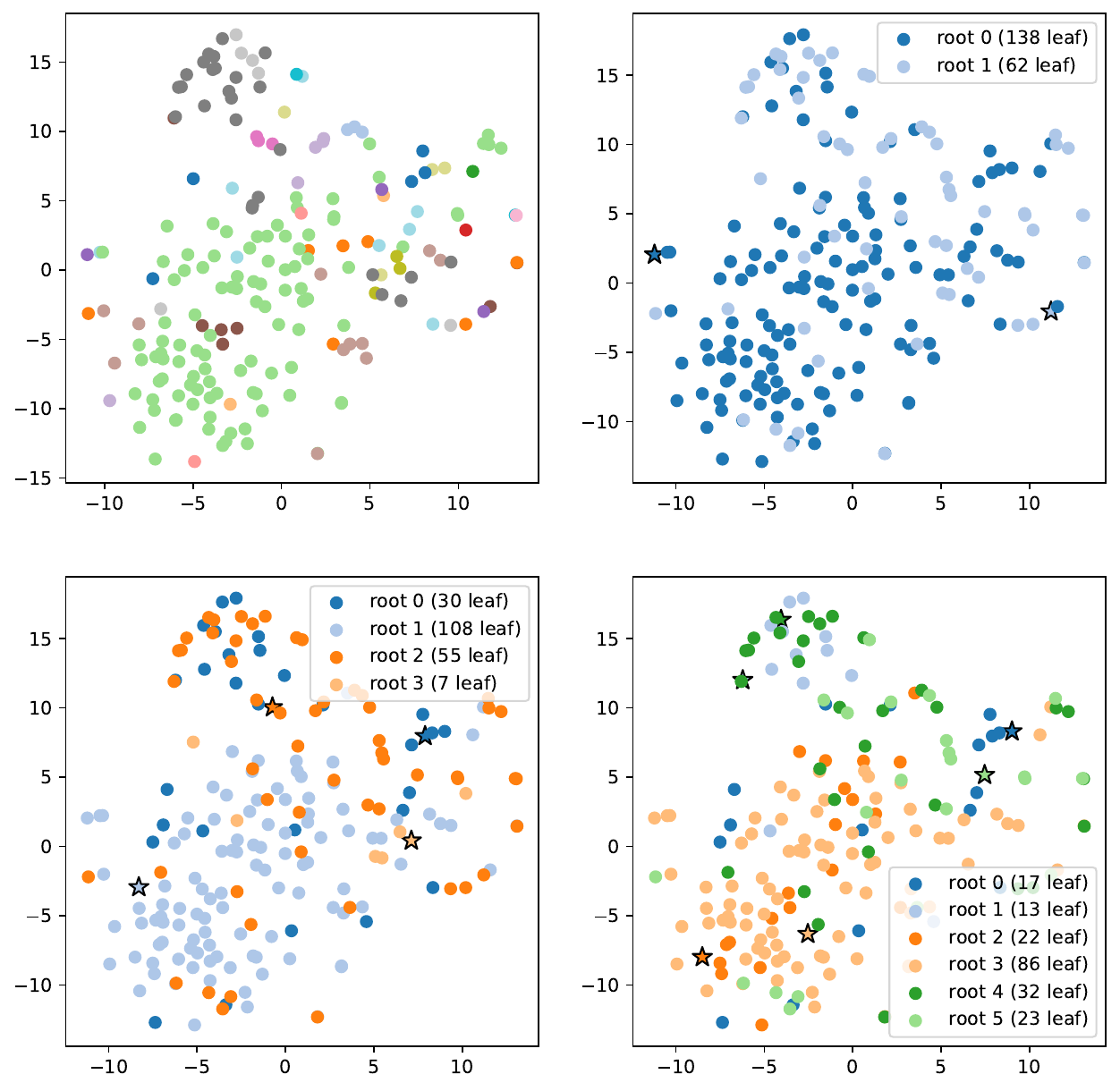}
 \caption{
 (Top) Ensembling effect of VANNS in MD3 task.
 (Bottom 4)
 The ground-truth cluster label (top-left) and
 the nodes of the nearest-neighbor search tree found by VANNTEC (Qwen3-32B, Medrxiv)
 at depth 1-3 colored by the shared parents ($\star$),
 plotted with t-SNE embeddings of all-MiniLM-L6-v2.
 VANNTEC's explicit search tree provides high interpretability.
 }
 \label{fig:clustering-misc}
\end{figure}

\begin{figure}[p]
 \centering
 \begin{minipage}{.9\linewidth}
  \includegraphics[width=\linewidth]{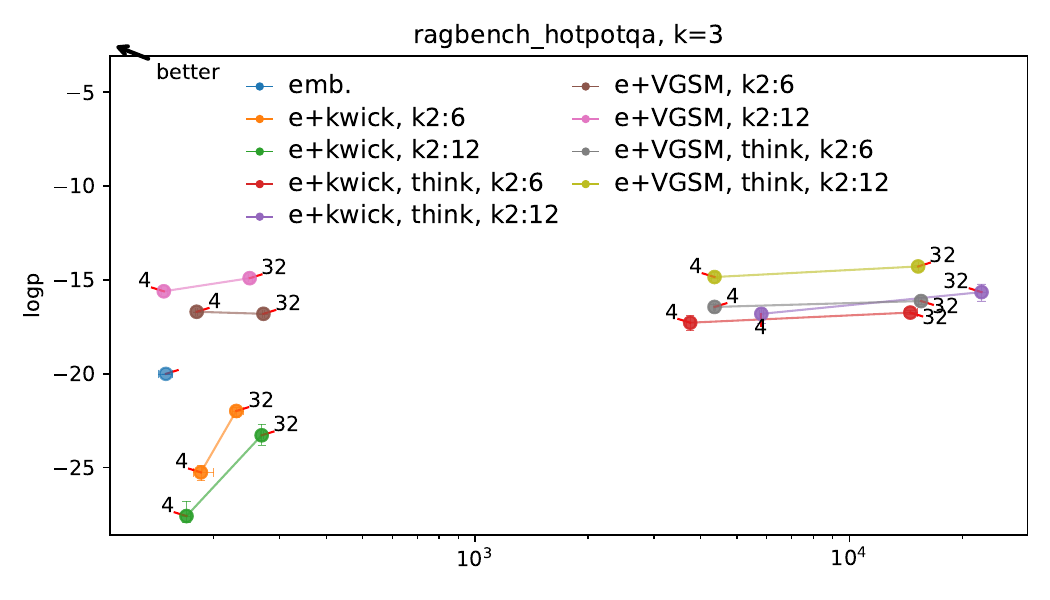}
  \includegraphics[width=\linewidth]{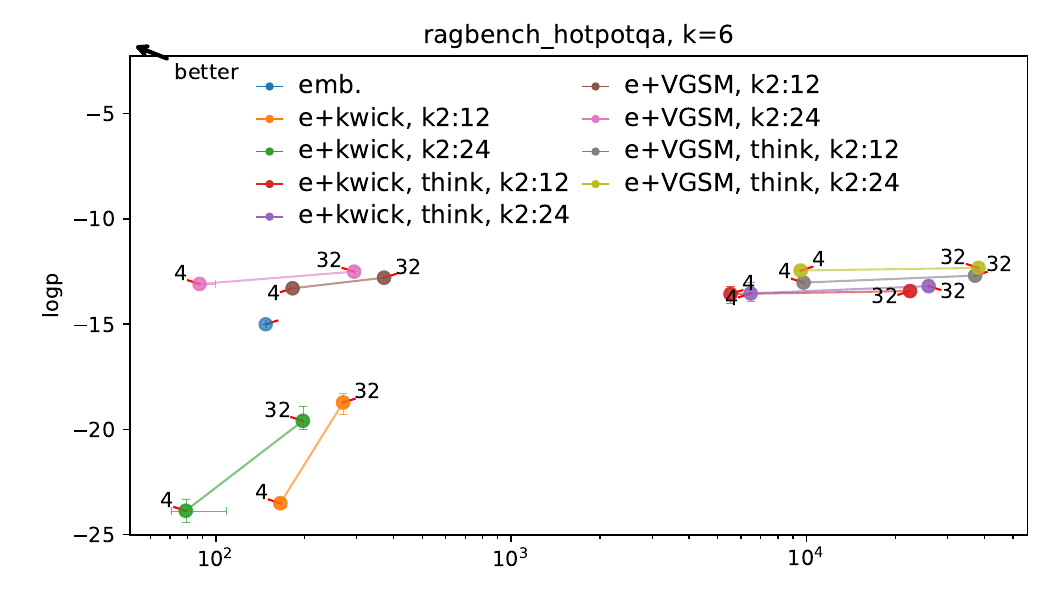}
  \includegraphics[width=\linewidth]{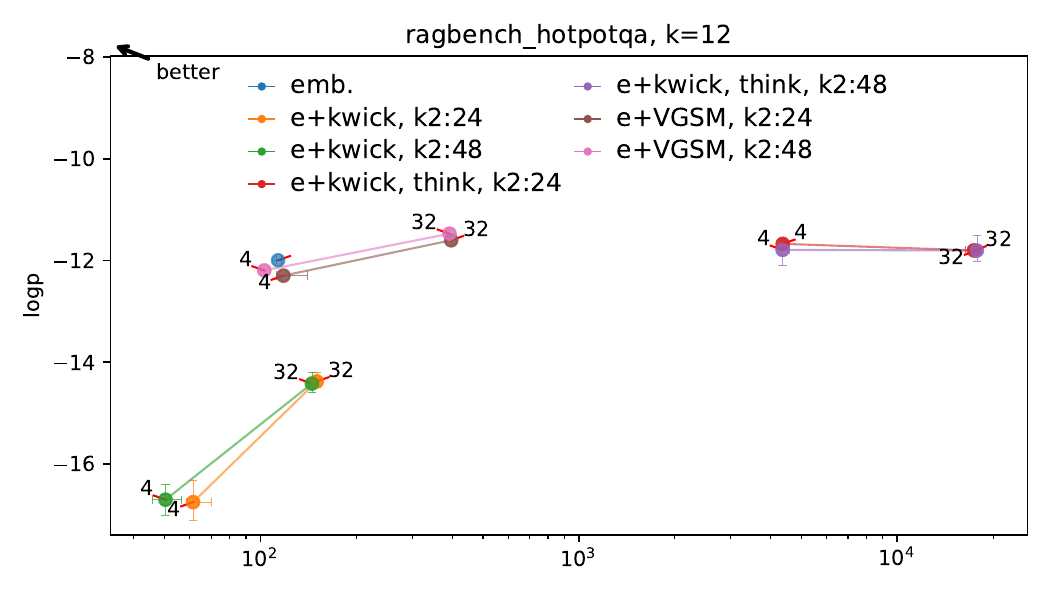}
 \end{minipage}
\caption{
 The speed-accuracy pareto front
 ($\log x$-axis: the runtime per query, $y$-axis: \function{logp})
 of RAG tasks (HotpotQA)
 over two Qwen3 model sizes (4B, 32B).
}
\label{fig:rag-logp-hotpotqa}
\end{figure}

\begin{figure}[p]
 \centering
 \begin{minipage}{.9\linewidth}
  \includegraphics[width=\linewidth]{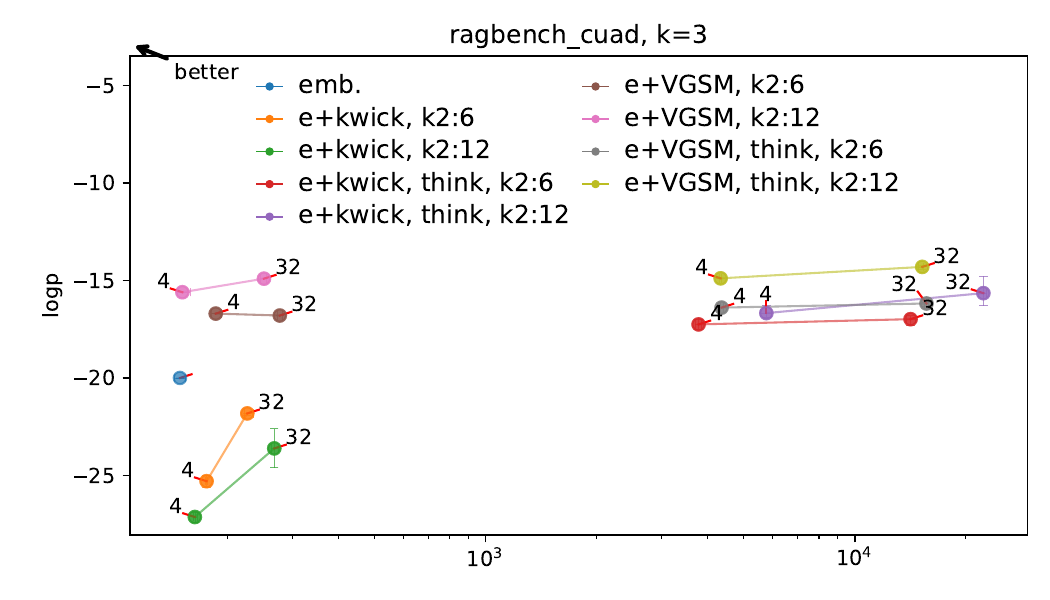}
  \includegraphics[width=\linewidth]{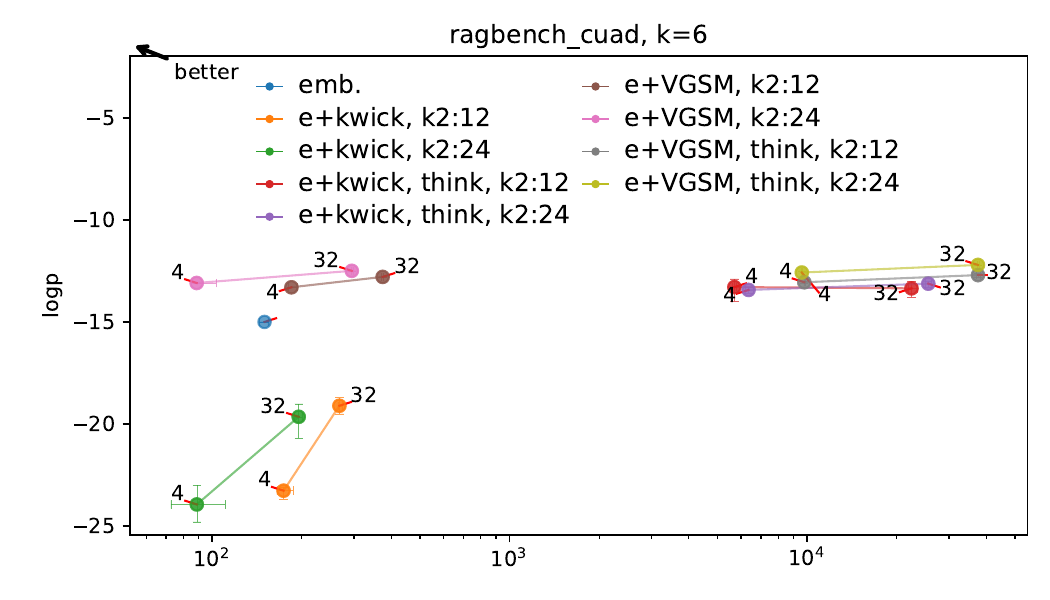}
  \includegraphics[width=\linewidth]{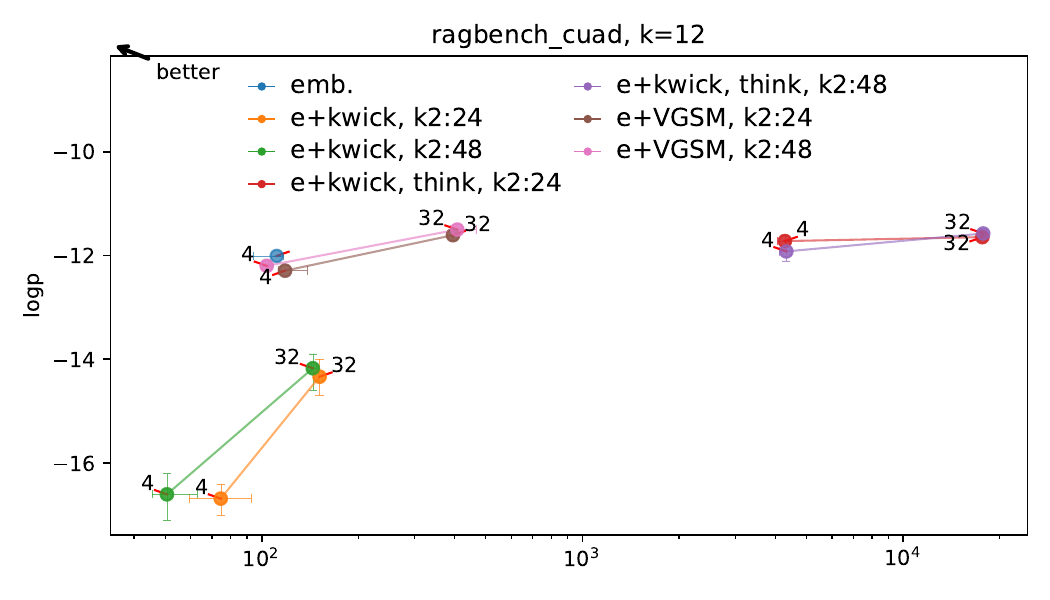}
 \end{minipage}
\caption{
 The speed-accuracy pareto front
 ($\log x$-axis: the runtime per query, $y$-axis: \function{logp})
 of RAG tasks (CUAD)
 over two Qwen3 model sizes (4B, 32B).
}
\label{fig:rag-logp-cuad}
\end{figure}

\begin{figure}[p]
\centering
 \begin{minipage}{.9\linewidth}
 \includegraphics[width=\linewidth]{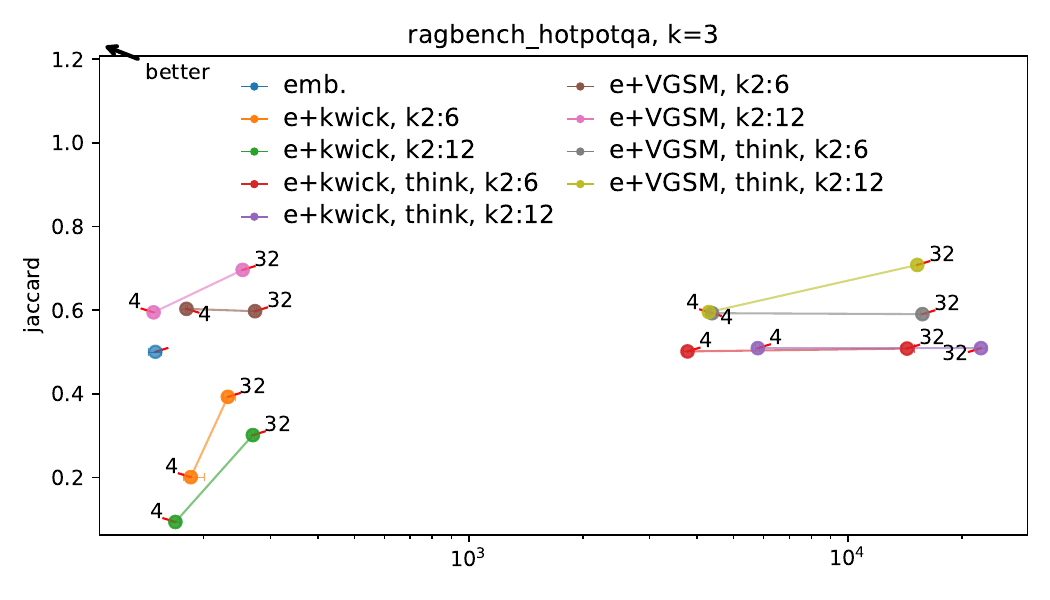}
 \includegraphics[width=\linewidth]{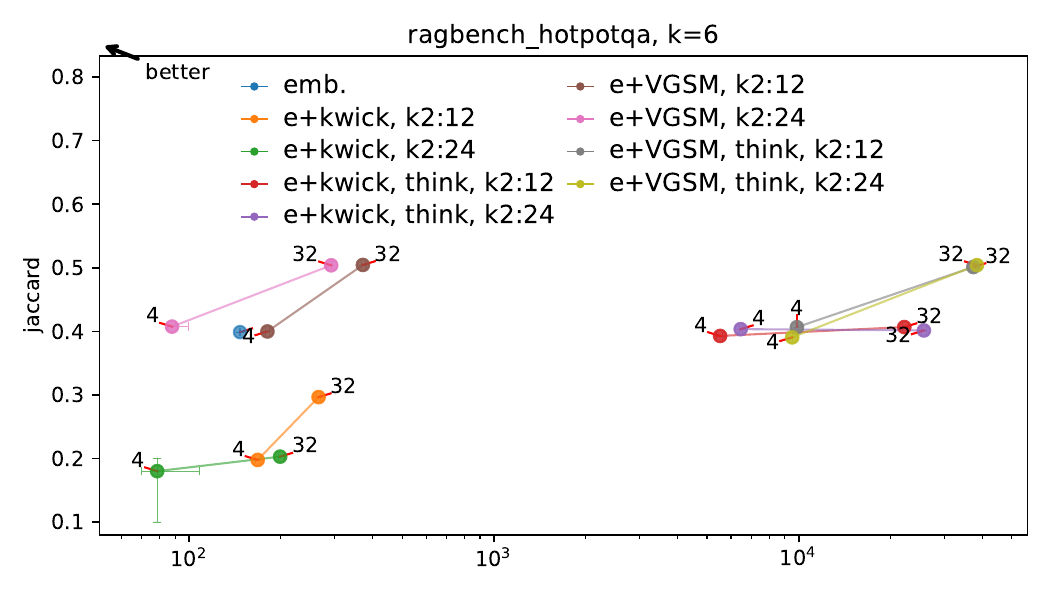}
 \includegraphics[width=\linewidth]{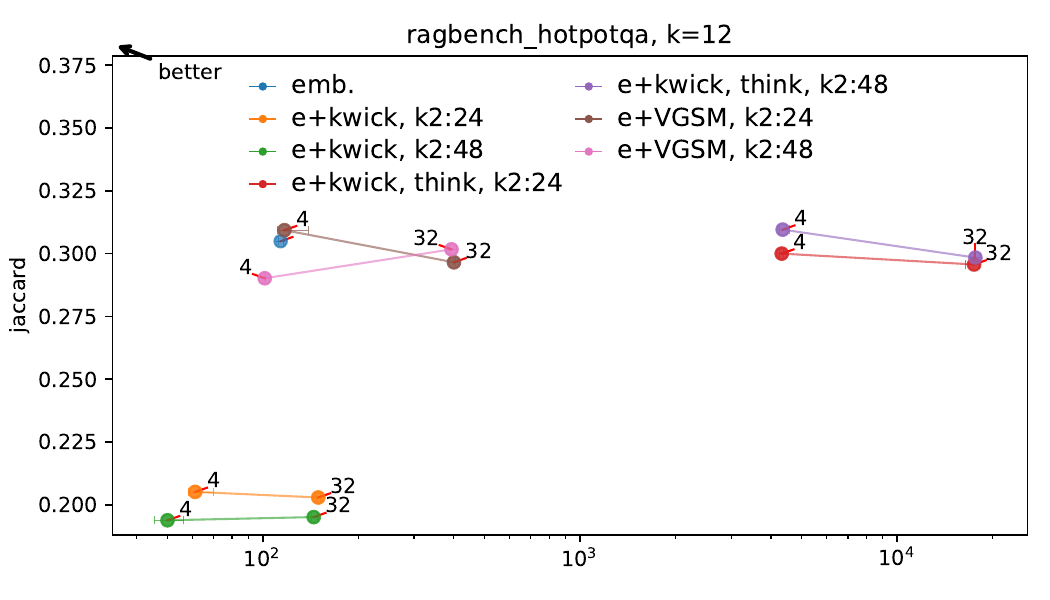}
 \end{minipage}
\caption{
 The speed-accuracy pareto front
 ($\log x$-axis: the runtime per query, $y$-axis: Jaccard distance
 $\frac{|S\cap S^*|}{|S\cup S^*|}$ between the selected documents $S$ and the ground truth set $S^*$)
 of RAG tasks (HotpotQA)
 over two Qwen3 model sizes (4B, 32B).
}
\label{fig:rag-jaccard-hotpotqa}
\end{figure}

\begin{figure}[p]
\centering
 \begin{minipage}{.9\linewidth}
 \includegraphics[width=\linewidth]{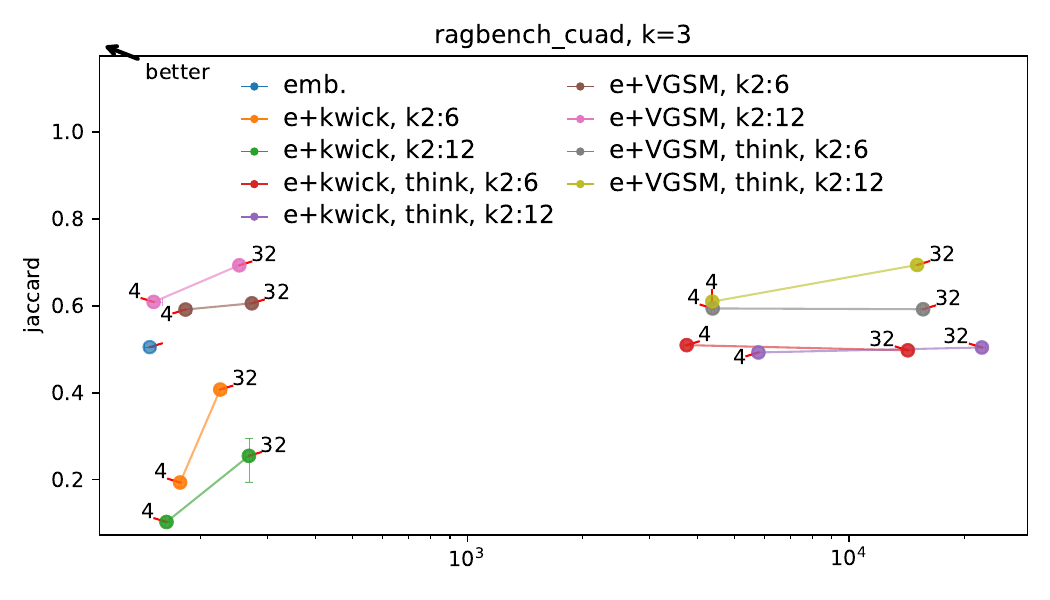}
 \includegraphics[width=\linewidth]{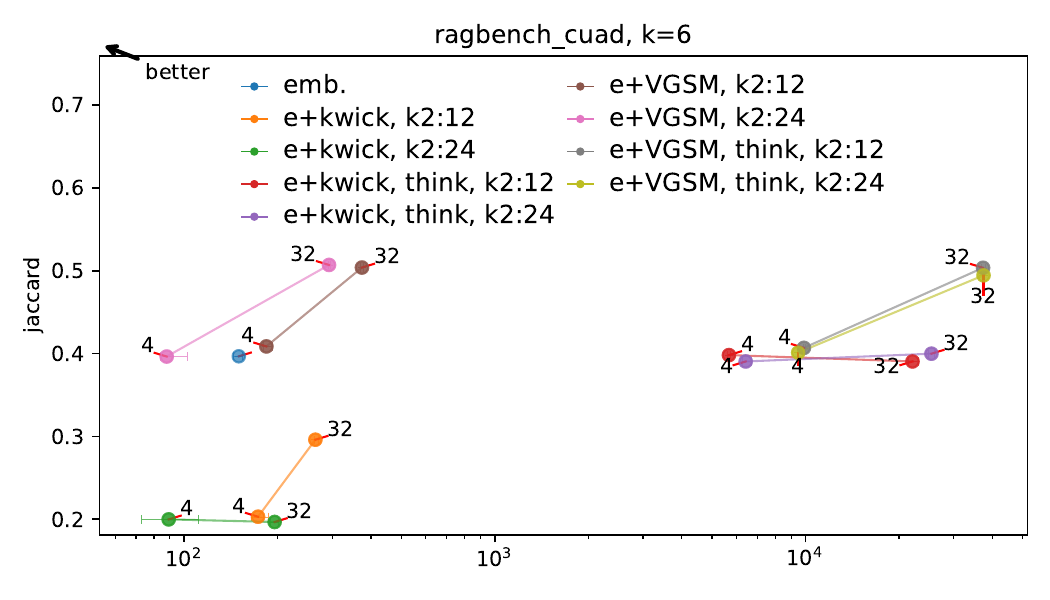}
 \includegraphics[width=\linewidth]{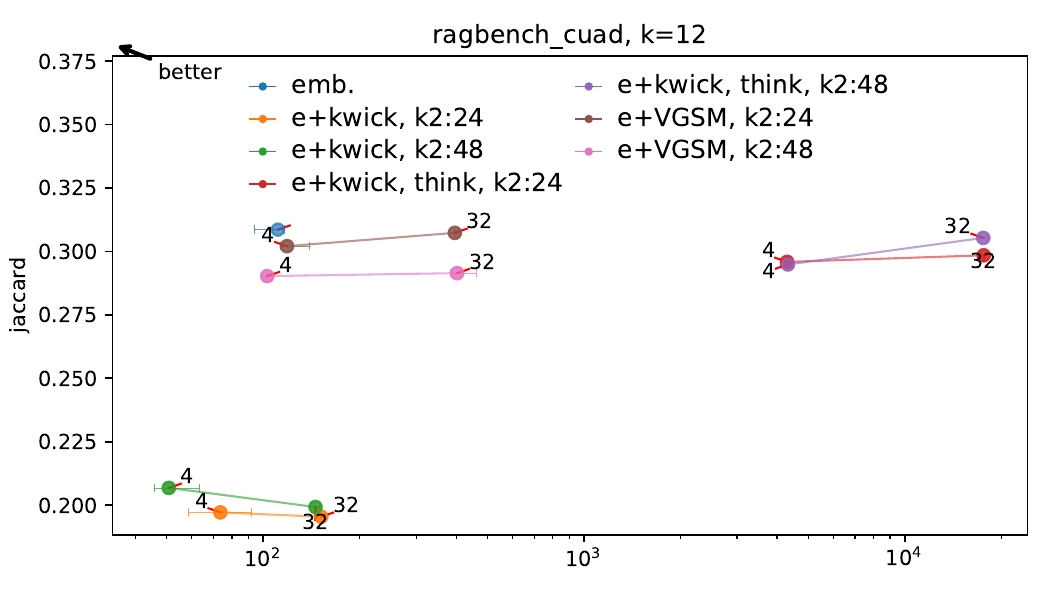}
 \end{minipage}
\caption{
 The speed-accuracy pareto front
 ($\log x$-axis: the runtime per query, $y$-axis: Jaccard distance
 $\frac{|S\cap S^*|}{|S\cup S^*|}$ between the selected documents $S$ and the ground truth set $S^*$)
 of RAG tasks (CUAD)
 over two Qwen3 model sizes (4B, 32B).
}
\label{fig:rag-jaccard-cuad}
\end{figure}

\begin{figure}[p]
 \hfill
 \begin{minipage}{0.48\linewidth}
  \includegraphics[width=\linewidth]{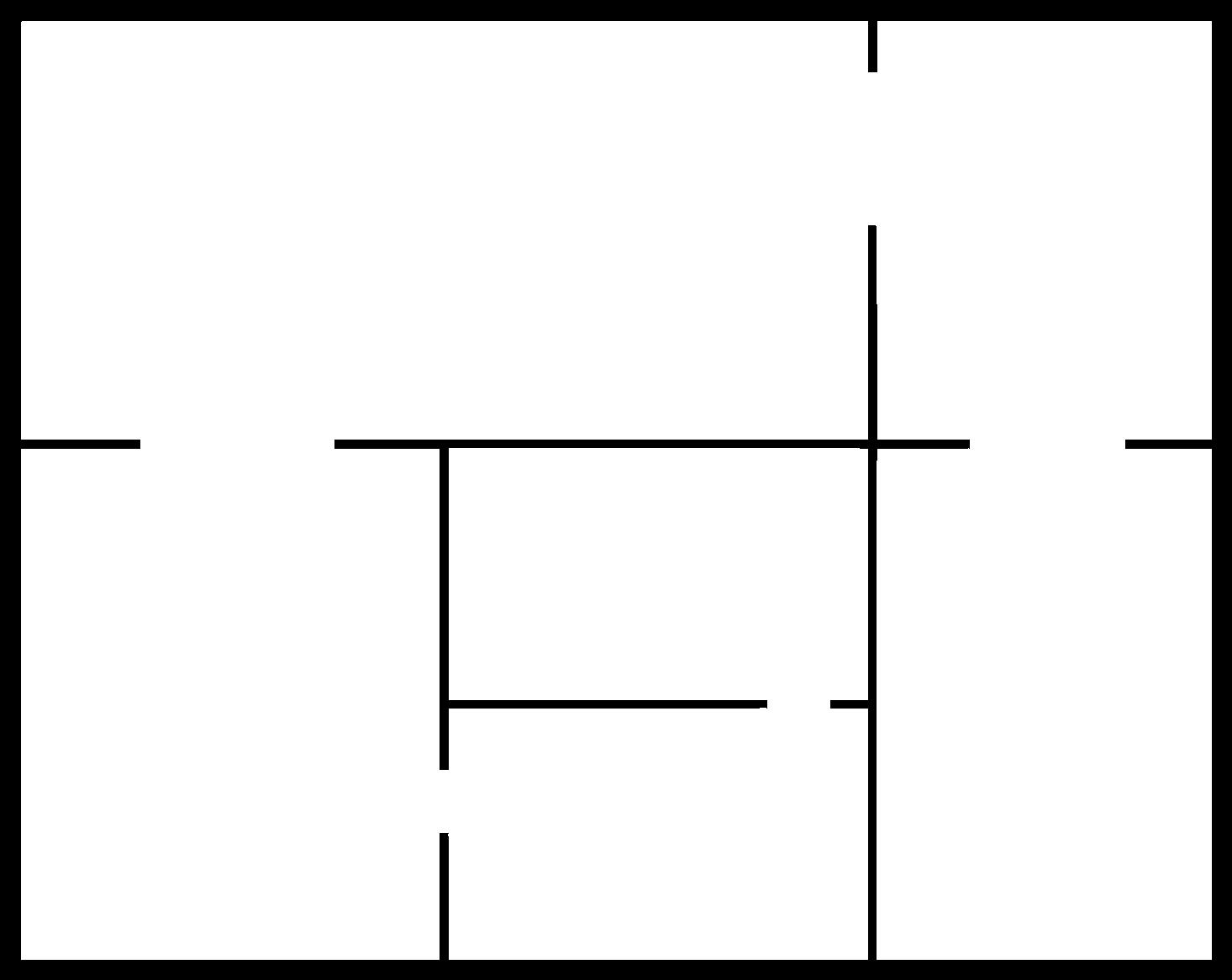}
 \end{minipage}
 \hfill
 \begin{minipage}{0.3\linewidth}
\begin{verbatim}
####################
#.............#....#
#..................#
#.............#....#
#......A......#....#
#.............#....#
#.............#....#
###..###########..##
#......#......#....#
#......#......#.A..#
#......#......#....#
#...A..########....#
#......#......#....#
#......#......#....#
#......#......#....#
\end{verbatim}
 \end{minipage}
 \hfill
 \caption{
 An example floor map obtained from HouseExpo dataset, and the optimal access point locations
 obtained by branch-and-bound using upper bounds that assume a modular objective.
 }
 \label{fig:wifi-floor}
\end{figure}

\end{document}